\documentclass[11pt]{article}

\usepackage[final]{acl}

\usepackage{times}
\usepackage{latexsym}

\usepackage[T1]{fontenc}
\usepackage[utf8]{inputenc}

\usepackage{microtype}

\usepackage{inconsolata}

\usepackage{graphicx}
\graphicspath{{figures/}}

\usepackage{multirow}
\usepackage{booktabs}

\usepackage{hyperref}
\usepackage{amsmath}
\usepackage{algorithm}
\usepackage{algorithmic}
\usepackage{enumitem}
\usepackage{xcolor}
\usepackage{graphicx}
\usepackage{subcaption}
\usepackage{diagbox} 
\usepackage{makecell}
\usepackage{array}

\usepackage{amssymb}

\usepackage{listings}
\usepackage[most]{tcolorbox}
\tcbuselibrary{breakable}
\usepackage{fvextra}
\usepackage{fontawesome5}

\definecolor{codegreen}{rgb}{0,0.6,0}
\definecolor{codegray}{rgb}{0.5,0.5,0.5}
\definecolor{codepurple}{rgb}{0.58,0,0.82}
\definecolor{backorange}{RGB}{255,250,240}
\definecolor{frameorange}{RGB}{255,140,0}
\definecolor{codebg}{rgb}{0.95,0.95,0.95}

\newcommand{\eg}{\emph{e.g.}}
\newcommand{\ie}{\emph{i.e.}}

\newcommand{\dataset}{IBCBench}
\newcommand{\model}{BundleWeaver}

\title{Weaving Visual Narratives: Agentic Image Bundle Composition Beyond Atomic Visual Matching}

\newcounter{corrauth}

\author{
  \textbf{Rong Shan\textsuperscript{1,2}},
  \textbf{Tianyi Xu\textsuperscript{1}},
  \textbf{Congmin Zheng\textsuperscript{1}},
  \textbf{Wenteng Chen\textsuperscript{1}},
  \\
  \textbf{Jiachen Zhu\textsuperscript{1}},
  \textbf{Junjie Wu\textsuperscript{3}},
  \textbf{Dun Zeng\textsuperscript{3}},
  \textbf{Teng Wang\textsuperscript{3}},
  \\
  \textbf{Weiwen Liu\textsuperscript{1}},
  \textbf{Changwang Zhang\textsuperscript{3}},
  \textbf{Weinan Zhang\textsuperscript{1}},
  \textbf{Jun Wang\textsuperscript{3}\thanks{Corresponding authors}}
  \setcounter{corrauth}{\value{footnote}},
  \textbf{Jianghao Lin\textsuperscript{1}}
  \footnotemark[\value{corrauth}]
  \\
  \textsuperscript{1} Shanghai Jiao Tong University,
  \textsuperscript{2} Shanghai Innovation Institute,
  \textsuperscript{3} OPPO
  \\
  \texttt{shanrong@sjtu.edu.cn, linjianghao@sjtu.edu.cn}
  \\
{
    \href{https://github.com/LaVieEnRose365/Image-Bundle-Composition}{
      \faGithub \ GitHub
    }
    \quad
    \href{https://huggingface.co/datasets/CyberDancer/IBCBench}{
      \raisebox{-0.18em}{
        \includegraphics[height=1.25em]{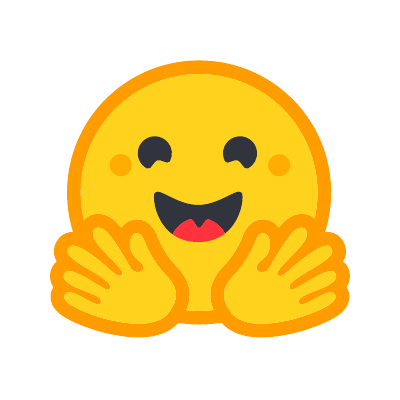}
      }
      Hugging Face
    }
  }
}

\begin{document}
\maketitle
\begin{abstract}
Image retrieval has traditionally been formulated as a point-wise matching problem, where each candidate image is scored in isolation. However, this atomic paradigm fails to capture the complexity of human search intent within personal photo collections, where users often seek compact visual stories bound by structural relations rather than isolated snapshots. To address this limitation, we introduce \textbf{Image Bundle Composition (IBC)}, a novel paradigm that shifts the objective from ranking individual images to dynamically composing cohesive image bundles from a massive, unstructured photo pool. Since target bundles are not predefined, IBC presents a severe combinatorial explosion challenge and demands modeling non-decomposable joint relevance. To establish this paradigm, we construct \dataset, the first IBC benchmark dataset containing 109,467 images and 667 verified queries, built via a semi-automated verification pipeline. Furthermore, we propose \textbf{\model} , an agentic framework that reformulates IBC as query-conditioned incremental hyperedge discovery. By employing a Large Language Model to adaptively search for missing relational roles and utilizing a Vision-Language Model for whole-bundle verification, \model\ effectively navigates the combinatorial space. Extensive experiments demonstrate that while state-of-the-art embedding models and static decompose-and-rerank paradigms suffer from relational blindness, \model\ achieves substantial performance gains, highlighting the necessity of shifting from atomic scoring to dynamic relational composition. Our dataset and code are available.
\end{abstract}

\section{Introduction}

\begin{figure*}[t]
    \centering
    \vspace{-7pt}
    \includegraphics[width=0.97\textwidth]{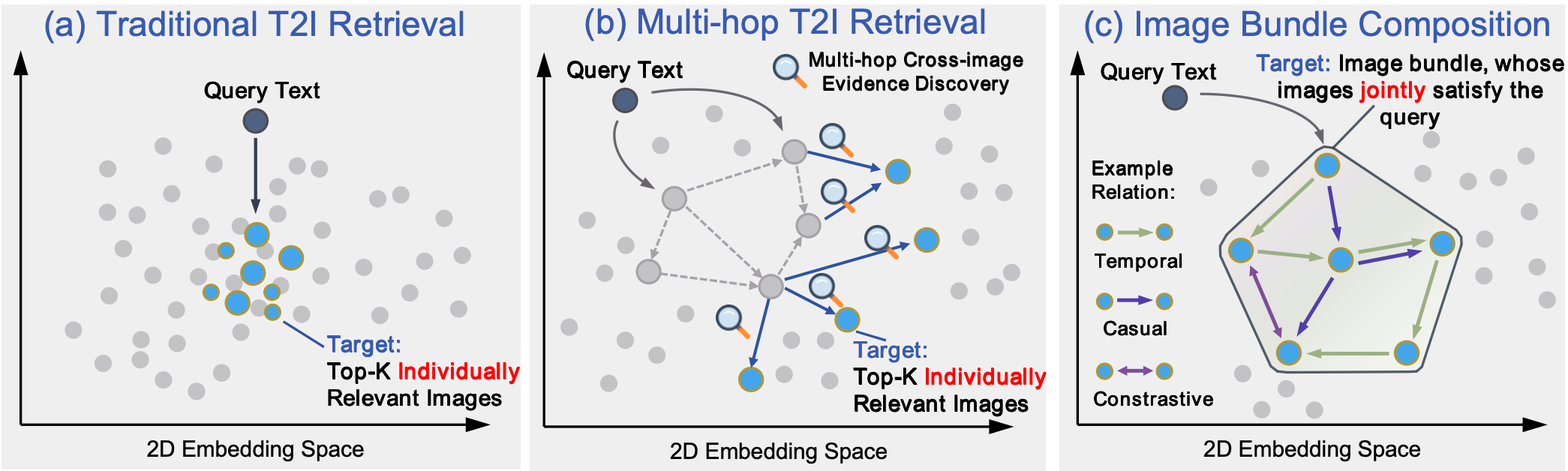}
    \vspace{-3pt}
    \caption{Comparison of different text-to-image retrieval paradigms.}
    \vspace{-13pt}
    \label{fig: intro}
\end{figure*}

% Image retrieval has traditionally been formulated as a point-wise matching problem between a textual query and individual visual assets~\cite{li2026qwen3, xu2026photobench, deng2026deepimagesearch}. 
% With the rapid development of embedding models, modern retrieval systems have achieved remarkable success in aligning natural language with visual content~\cite{xue2025improve, jiang2024vlm2vec}. 
% Despite this progress, the dominant paradigm remains fundamentally atomic: given a query, each candidate image is scored in isolation, yielding a ranked list of independently relevant items.

Image retrieval has traditionally been formulated as a point-wise matching problem between a textual query and individual visual assets~\cite{li2026qwen3, xu2026photobench, deng2026deepimagesearch}. Despite the progress of multimodal embedding models~\cite{xue2025improve, jiang2024vlm2vec}, the dominant paradigm remains fundamentally atomic: given a query, each candidate image is scored in isolation, yielding a ranked list of independently relevant items~\cite{lin2024rella, shan2025automatic}.

As illustrated in Figure~\ref{fig: intro}(a), standard evaluation on benchmarks such as MSCOCO~\cite{lin2014microsoft}, Flickr30K~\cite{plummer2015flickr30k}, and MMEB~\cite{jiang2024vlm2vec} operate strictly under this atomic assumption. 
Recently, some approaches~\cite{deng2026deepimagesearch} have extended this paradigm to multi-hop text-to-image retrieval (Figure~\ref{fig: intro}(b)), navigating through the image pool via cross-image evidence discovery. 
Yet, despite incorporating intermediate visual reasoning, the ultimate retrieval target is still a ranked list of individually scored, disconnected images.

However, this atomic paradigm fails to capture the complexity of human memory and search intent. 
In real-world scenarios, users rarely recall experiences as isolated snapshots, rather, they remember compact visual stories~\cite{andrews2011clues, kurby2008segmentation}. 
A user might search for the key highlights of a concert night, the gradual transition of a landscape from day to dusk, or a continuous journey spanning multiple landmarks. 
In such cases, the ideal target is not a single optimal image, nor a disjointed set of individually relevant results, but a cohesive \textit{image bundle} whose members jointly satisfy the query. 
While the individual images within a bundle may be visually heterogeneous, they are bound together by an underlying relation, \eg, temporal progression, event summarization, or spatial continuity, which renders them meaningful only as a collective whole.

Motivated by this observation, we introduce \textbf{Image Bundle Composition (IBC)}, a novel paradigm that shifts the retrieval objective from ranking individual images to dynamically composing cohesive image bundles from a vast, unstructured photo pool. 
Crucially, these bundles are \textit{not} pre-defined or indexed a priori. As any arbitrary subset of images could theoretically form a bundle, the search space poses a challenge of  combinatorial explosion. 
Therefore, IBC demands more than just matching a description to an image. It requires the system to actively navigate a massive combinatorial space to discover and compose a unique combination of images that collectively instantiates the intended narrative or relational structure.

Establishing this new paradigm necessitates a dedicated benchmark, yet creating a dataset for IBC presents significant challenges. 
Exhaustive subset annotation is computationally impossible due to the combinatorial nature of the task, while unconstrained discovery using Vision-Language Models (VLMs) often yields bundles that lack uniqueness or structural coherence. 
To tackle this, we design a semi-automated mining and verification pipeline. 
By anchoring the search space to spatiotemporal sessions, we extract candidate sliding windows and employ a VLM to rigorously evaluate their cross-image relations. 
Following this, human annotators refine the VLM-generated queries and filter out ambiguous cases, ensuring that each ground-truth bundle represents a unique, unambiguous answer within the global image pool. Finally, the resulting \textbf{\dataset} consists of  109,467 images and 667 meticulously verified queries.

To tackle the combinatorial search space of IBC, we propose \textbf{\model}, an  agentic framework that reformulates the task as query-conditioned hyperedge discovery. 
Viewing the unstructured image corpus as a graph of vertices, a target bundle represents a hyperedge connecting a specific subset through a higher-order semantic relation. 
To avoid exhaustive enumeration, \model\ explores this space incrementally. 
Starting from diverse seed images, it expands partial bundles by employing a LLM reasoning agent to deduce adaptive search directions, conditioned on both the original query and the current bundle state. 
Instead of greedily retrieving more locally relevant images, our method deliberately searches for images that fulfill \textit{missing roles} within the ongoing visual composition. 
Supported by adaptive beam search, contextual candidate pruning, and a whole-bundle VLM reranker, \model\  effectively identifies promising cohesive sets. 
Empirical results on IBCBench demonstrate that while current state-of-the-art multimodal embeddings and VLMs excel at finding isolated matches, they struggle to compose them logically. 
\model\ bridges this gap, offering substantial performance gains by explicitly modeling and searching for cross-image compositions.

Our contributions are summarized as follows:
\begin{itemize}[leftmargin=10pt]
    \item We formulate \textbf{Image Bundle Composition} (IBC), a novel retrieval paradigm that shifts the focus from isolated image matching to the dynamic composition of coherent image bundles.
    
    \item We construct and open-source the first IBC benchmark  \textbf{IBCBench}, comprising 109,467 images and 667 meticulously verified queries, built via a robust semi-automated pipeline.
    
    \item We further propose \textbf{\model}, an innovative agentic framework that models IBC as query-conditioned hyperedge discovery. By incrementally fulfilling missing bundle roles and employing whole-bundle reranking, \model\ demonstrates promising results and sets a strong baseline for the task.
\end{itemize}
\section{Task Formulation}

To elucidate the fundamental differences between our proposed task and existing paradigms, we first formalize the traditional image retrieval pipeline, and then introduce the mathematical formulation of Image Bundle Composition (IBC).

\subsection{Preliminaries: Traditional Text-to-Image Retrieval}
Let $\mathcal{I} = \{x_1, x_2, \dots, x_N\}$ denote a large, unstructured image pool of size $N$, and $q$ denote a natural language query. 
In traditional T2I retrieval, the objective is formulated as a point-wise matching problem. 
The system defines a scoring function $f(x, q)$ that independently measures the semantic similarity between the query and each individual candidate image $x \in \mathcal{I}$. 
The goal is to retrieve an image $x^*$ (or a ranked list of top-$k$ images) that maximizes this independent relevance score:
\begin{equation}
    x^* = \arg\max_{x \in \mathcal{I}} f(x, q).
\end{equation}
In this paradigm, the relevance of any image is atomic and entirely decoupled from the presence or absence of other images in the retrieved list.

\subsection{Image Bundle Composition (IBC)}
Unlike traditional retrieval, IBC does not seek a single optimal image or a disconnected list of items. Instead, it aims to dynamically compose a cohesive \textit{image bundle}, \ie, a compact subset of images, that collectively fulfills the user's relational or narrative intent.

Formally, let $B \subseteq \mathcal{I}$ denote a candidate image bundle consisting of $K$ distinct images, where the bundle size $K = |B|$ is dynamically determined but bounded by a small integer $K_{max}$ (\eg, $2 \le K \le K_{max}$).
The objective of IBC is to discover the optimal subset $B^*$ that maximizes a \textit{joint} relevance scoring function $\Phi(B, q)$:
\begin{equation}
\label{eq: BIR objective}
    B^* = \arg\max_{B \subseteq \mathcal{I}, 2 \le |B| \le K_{max}} \Phi(B, q),
\end{equation}
where $\Phi(\cdot, \cdot)$ evaluates the cross-image composition as a whole. Rather than measuring independent visual-textual alignment, $\Phi$ captures the structural, temporal, or spatial relations dictated by the query $q$ across all elements in $B$.

\subsection{Discussions on Challenges of IBC}
\label{sec: challenge dicussion}

IBC introduces two profound mathematical and computational challenges that fundamentally distinguish it from standard retrieval:
\begin{itemize}[leftmargin=10pt]
    \item \textbf{Combinatorial Explosion:} 
Because bundles are \textit{not predefined or indexed a priori}, the system must actively search over the subsets of $\mathcal{I}$. The size of this search space, denoted as $\mathcal{S}$, scales combinatorially with the pool size $N$:
\begin{equation}
    |\mathcal{S}| = \sum_{k=2}^{K_{max}} \binom{N}{k} \approx \mathcal{O}(N^{K_{max}}).
\end{equation}
For a typical personal photo collection where $N \sim 10^4$, evaluating all candidate combinations exhaustively is computationally intractable, making brute-force subset selection impossible.

\item \textbf{Non-Decomposability of Joint Relevance:}
Crucially, the joint relevance function $\Phi(B, q)$ is non-decomposable. This means the bundle-level score cannot be approximated by aggregating individual image-level scores:
\begin{equation}
    \Phi(B, q) \neq g \Big( \{ f(x_i, q) \mid x_i \in B \} \Big),
\end{equation}
where $g$ is any monotonic aggregation operator (\eg, sum or average). 
For instance, if $q$ asks for \textit{the transition of the Eiffel Tower from day to night}, two visually stunning daytime photos of the tower might each receive high independent scores $f(x, q)$. However, their combination fails to satisfy the \textit{transition} relation, yielding a low joint score $\Phi(B, q)$. 
Consequently, a greedy strategy that simply selects the top-$K$ independently retrieved images will inherently fail in IBC, necessitating a paradigm shift from independent scoring to relational composition.

\end{itemize}
We provide more detailed  theoretical analysis on limitations of atomic retrieval in IBC in Appendix~\ref{appendix:appendix_theoretical_limitations}.

% \section{Dataset Construction}

\begin{figure*}[t]
    \centering
    \vspace{-7pt}
    \includegraphics[width=0.99\linewidth]{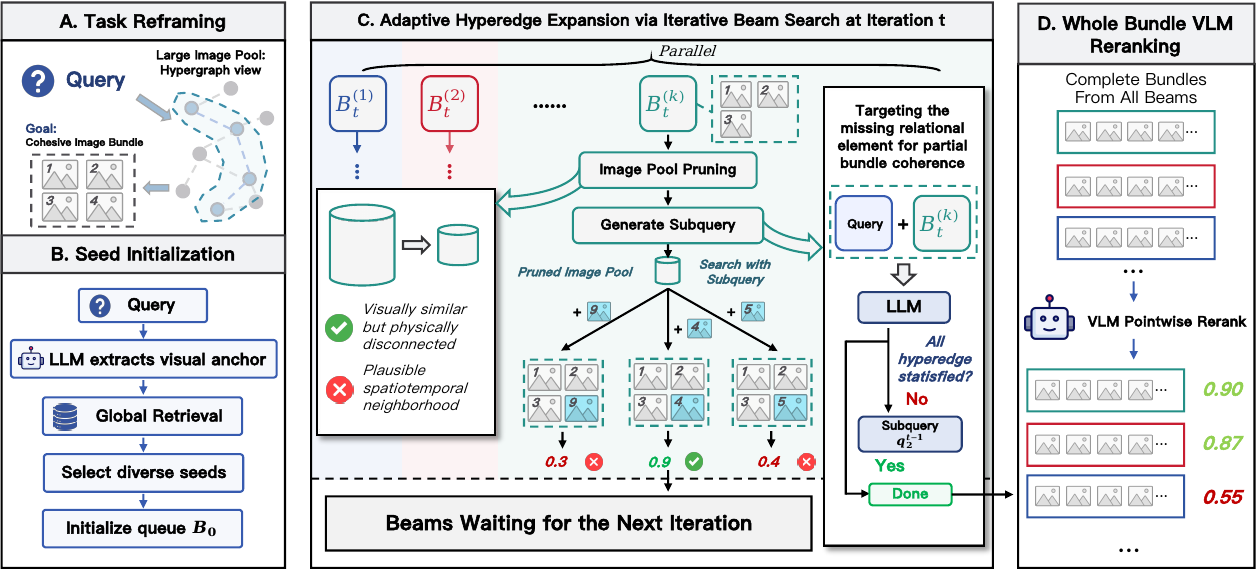}
    % \vspace{-3pt}
    \caption{Framework illustration of our proposed BundleWeaver.}
    \vspace{-5pt}
    \label{fig: main}
\end{figure*}

\section{Dataset Construction}
\label{sec:dataset_construction}

Constructing a dataset for IBC is highly challenging. Exhaustive manual annotation is computationally impossible due to combinatorial explosion, while unconstrained VLM generation often yields ambiguous or decomposable image lists. To fill this blank, we introduce \textbf{\dataset}, the first dataset dedicated to IBC. 
As illustrated in Figure~\ref{fig: dataset construction}, 
We design a five-stage semi-automated pipeline. Due to space constraints, we outline the core principles here and defer more details to Appendix~\ref{appendix: dataset construction}.

\paragraph{Property Requirements.} A valid candidate subset $B \subset \mathcal{I}$ must satisfy four strict criteria to be accepted: \textbf{(C1) Joint Completeness:} Every image plays an indispensable role; \textbf{(C2) Cross-image Binding:} The query enforces a higher-order relation (\eg, temporal progression) rather than merely listing independent visual targets; \textbf{(C3) Uniqueness:} The bundle should be unambiguous, and alternatives should be minimized; \textbf{(C4) Bounded Redundancy:} Images within $B$ must be visually distinct.

\paragraph{Candidate Mining.} We source our raw pool from the YFCC-100M dataset~\cite{thomee2016yfcc100m}, retaining 109,467 photos with valid spatiotemporal metadata. We pre-extract dense captions and tags using GPT-4o and compute multimodal embeddings to penalize redundancy. To bypass the $\mathcal{O}(N^K)$ search space, we group user photos into spatiotemporal sessions and extract sliding windows of size $K \in [3, 5]$. A composite heuristic score aggressively prunes these windows based on spatiotemporal coverage and semantic richness, yielding 7,460 highly diverse candidate windows.

\paragraph{VLM Verification \& Human Review.} We employ Claude-Opus-4.5~\cite{anthropic2025claudeopus45} as a rigorous semantic verifier to check candidates against predefined relational templates (\ie, \textit{Same-location Dynamics} and \textit{Cross-location Structures}). The VLM must generate a natural query and articulate a specific cross-image shared anchor. Finally, four expert annotators carefully review the VLM-approved candidates against C1-C4, filtering out ambiguous cases to ensure global uniqueness.

\paragraph{Final Dataset Statistics.} Through this exhaustive pipeline (with an end-to-end acceptance rate $< 9\%$), the final \dataset\ dataset comprises 667 high-quality, verified queries. Bundle sizes are distributed across 3 (24.3\%), 4 (32.2\%), and 5 (43.5\%) images. The semantic relations are well-balanced, with 52.5\% focusing on same-location dynamics and 47.5\% capturing cross-location structural ties.

\section{Methodology}

Addressing IBC by exhaustively evaluating all possible image subsets is computationally intractable. Furthermore, because bundles are dynamically defined by the user's relational intent rather than indexed a priori, traditional atomic retrieval mechanisms cannot be directly applied. To resolve these challenges, we propose \textbf{\model}, an agentic framework that reformulates IBC as a problem of \textit{query-conditioned hyperedge discovery} and solves it through an adaptive, incremental search process.

As is shown in Figure~\ref{fig: main}, the \model\ framework consists of three core stages, \ie, Diverse Seed Initialization, Adaptive Hyperedge Expansion via Beam Search, and Whole-Bundle VLM Reranking.

\subsection{Reformulation: Incremental Hyperedge Discovery}
Let the massive image pool $\mathcal{I}$ be the vertex set $\mathcal{V}$ of an implicit, fully connected hypergraph. A coherent image bundle $B$ corresponds to a hyperedge connecting a small subset of vertices. Since the set of valid hyperedges $\mathcal{E}$ is not predefined, we can only dynamically \textit{construct} the hyperedges from scratch. \model\ models this construction as an incremental pathfinding problem. Starting from a single seed vertex (\ie, an image), the system sequentially discovers new vertices that fulfill missing narrative or relational roles, eventually closing the hyperedge.

\subsection{Diverse Seed Initialization}
The discovery process begins by anchoring the search space to a set of highly promising starting vertices. Given the original natural language query $q$, we prompt a LLM to extract the primary visual anchor and generate an initial search direction $d_1$. 
Using a dense multimodal embedding model, we encode $d_1$ and perform an initial global Nearest Neighbor search across the entire unstructured pool of $N$ images.

To prevent the search from collapsing into a single local optimum or retrieving near-duplicate starting points~\cite{xi2026survey, zhou2026externalization, zhu2025evolutionary}, we implement a \textbf{Diverse Seed Initialization} strategy. From the top-$K$ globally retrieved images, we select a set of $N_s$ seed images that exhibit maximum semantic and visual diversity (\eg, spanning different sub-clusters or contextual sessions). Each selected seed then acts as the independent root for a parallel hyperedge construction tree. This parallelized initialization ensures a broad coverage of candidate starting spaces across the global image pool without loss of generality.

\subsection{Adaptive Hyperedge Expansion via Parallel Beam Search}
Starting from the initialized diverse seeds, \model\ executes parallel branch expansions, searching for complementary images via a beam search of width $B_u$. At step $k$ within a specific search branch, let the current partial bundle be $B_{k-1} = \{x_1, x_2, \dots, x_{k-1}\}$. The expansion involves three key mechanisms:
\begin{itemize}[leftmargin=10pt]
    % \item \textbf{Metadata Pruning:} To restrict the branching factor of each parallel tree, we dynamically prune the vertex pool based on spatiotemporal constraints. For a given branch, we only consider candidate vertices $x_k$ that fall within a plausible spatiotemporal neighborhood relative to the seed image. This metadata gating slashes the active search space from $\sim 10^5$ to $\sim 10^2$ viable candidates per step.

% \item \textbf{Contextual Candidate Pruning:} 
% Rather than exposing the reasoning agent to the entire noisy pool, which is filled with visually similar but physically disconnected distractors, we bound the candidate expansions $x_k$ to a plausible spatiotemporal neighborhood relative to the seed image. This localized gating reduces the active hypothesis space from $\sim 10^5$ to $\sim 10^2$ viable candidates per step. Importantly, this step acts as a relational shield. It ensures that the subsequent semantic sub-query matching operates within a physically plausible boundary, preventing the beam search from being overly derailed by global noise.
\item \textbf{Contextual Candidate Pruning:}
Many IBC queries seek cohesive visual narratives grounded in real-world contexts, \eg, \textit{a trip to Florida}. Such queries naturally imply coarse temporal or geographic feasibility constraints. Therefore, rather than exposing the reasoning agent to the entire unconstrained pool, which is filled with visually similar but physically disconnected distractors, we allow it to dynamically bound the candidate expansions $x_k$ to a plausible spatiotemporal neighborhood relative to the seed image. However, this does not by itself solve the task, as images nearby in time or space may still be relationally incompatible. The following stages are therefore essential for composing a coherent bundle.

  % To restrict the branching factor of each parallel tree, we dynamically prune the vertex pool based on contextual constraints. For a given branch, we only consider candidate vertices $x_k$ that fall within a plausible spatiotemporal neighborhood relative to the seed image. This metadata gating slashes the active search space from $\sim 10^5$ to $\sim 10^2$ viable candidates per step.
% To restrict the branching factor of each parallel tree, we first constrain the search space to a compact contextual neighborhood around the seed image. For a given branch, we only consider candidate images $x_k$ that are likely to belong to the same underlying visual episode or trajectory, which keeps incremental expansion computationally tractable. Importantly, this step serves only as an efficiency filter, while scoring and selection is  driven by later stages. In practice, this localized gating reduces the active combinatorial space from $\sim 10^5$ images to $\sim 10^2$ viable candidates per step.

    \item \textbf{Adaptive Sub-query Generation:} Standard retrieval models lack the reasoning capability to deduce what is missing in a partial bundle. Therefore, we deploy an LLM as an active reasoning agent. Conditioned on the original query $q$ and the visual captions of the already selected images in $B_{k-1}$, the LLM dynamically generates an adaptive sub-query $d_k$ that specifically targets the missing relational element needed to advance the partial bundle coherence.

    \item \textbf{Path Scoring:} We encode the sub-query $d_k$ and retrieve the top-$C$ local candidates. To determine which branches survive in the beam search, we evaluate each candidate path using a composite scoring function that balances step-wise precision with whole-bundle coherence:
\begin{equation}
\resizebox{\columnwidth}{!}{$
\displaystyle
\text{Score}(B_k) =
\underbrace{
\frac{1}{k} \sum_{i=1}^{k} \cos(\mathbf{e}_{d_i}, \mathbf{e}_{x_i})
}_{\text{Step-wise Local Match}}
+ \lambda \cdot
\underbrace{
\cos\left(
\mathbf{e}_q,
\frac{\sum_{i=1}^{k} \mathbf{e}_{x_i}}
{\left\| \sum_{i=1}^{k} \mathbf{e}_{x_i} \right\|}
\right)
}_{\text{Holistic Bundle Alignment}}
$}
\label{eq:path scoring}
\end{equation}
where $\mathbf{e}_{d_i}$ and $\mathbf{e}_{x_i}$ are the dense embeddings of the $i$-th sub-query and selected image, respectively, and $\mathbf{e}_q$ is the embedding of the original query. $\lambda$ controls the trade-off between fulfilling individual sub-queries and maintaining the overall semantic trajectory of the hyperedge.
    
\end{itemize}

\subsection{Whole-Bundle VLM Reranking}
\label{sec: vlm rerank}

% Instead of rigidly expanding to a fixed depth, the parallelized beam searches adaptively terminate when the LLM reasoning agent determines that the bundle is complete and no further relational roles are required (bounded by a maximum depth $K_{max}$). Upon this dynamic termination, the process yields a rich, diverse set of completed bundle candidates of varying sizes. While the embedding-based path score effectively guides the heuristic search, dense embeddings often fail to capture fine-grained cross-image relations (\eg, verifying if the exact same person appears across all images, or maintaining strict identity consistency).

% To bridge this gap, we introduce \textbf{Whole-Bundle VLM Reranking} as the final selection mechanism. We merge the completed paths from all parallel search trees into a unified candidate pool. A Vision-Language Model acts as a pointwise reranker, simultaneously observing all images within a candidate bundle alongside the original query $q$. The VLM evaluates the structural coherence, uniqueness, and cross-image binding of each candidate, assigning a holistic score from 1 to 10. The highest-scoring hyperedge is returned as the final instantiated bundle $B^*$. This explicit decoupling of candidate generation (via adaptive search) and relational verification (via VLM reranking) effectively overcomes the combinatorial constraints while maximizing the relational quality of the final output.

Instead of expanding to a fixed depth, the beam search adaptively terminates when the LLM deems the bundle complete (bounded by a maximum depth $K_{max}$), yielding a diverse candidate pool. While embedding scores efficiently guide this heuristic search, they often fail to capture fine-grained cross-image relations like strict identity consistency. To this end, we propose \textbf{Whole-Bundle VLM Reranking}. We merge all completed paths into a unified pool, where a VLM acts as a pointwise reranker. By simultaneously evaluating all images within a candidate alongside the query $q$, the VLM scores its structural coherence and cross-image binding (1-10). The highest-scoring hyperedge is returned as $B^*$. This explicit decoupling of adaptive generation and relational verification efficiently mitigates combinatorial constraints while maximizing final output quality.

\begin{table*}[ht]

\centering

\vspace{-5pt}
\resizebox{0.99\textwidth}{!}{

\begin{tabular}{cc|cccc}
\toprule
\hline
\multicolumn{1}{c|}{Type} & Method                 & Precision & Recall & F1    & EM   \\ \hline
\multicolumn{1}{c|}{\multirow{4}{*}{\makecell{Multimodal \\ Embedding}}}              & CLIP-ViT-B/32           & 2.21            & 2.21            & 2.21           & 0.00             \\
\multicolumn{1}{c|}{}     & SigLIP2-giant          & 9.97      & 9.97   & 9.97  & 0.15 \\
\multicolumn{1}{c|}{}     & Qwen3-VL-Embed-8B  & 12.14     & 12.14  & 12.14 & 0.00 \\
\multicolumn{1}{c|}{}     & RzenEmbed              & 14.88     & 14.88  & 14.88 & 0.30 \\ \hline
\multicolumn{1}{c|}{\multirow{3}{*}{Caption + Text Embedding}}          & BM25                    & 3.29            & 3.29            & 3.29           & 0.00             \\
\multicolumn{1}{c|}{}     & BGE-M3                 & 7.05      & 7.05   & 7.05  & 0.15 \\
\multicolumn{1}{c|}{}     & Qwen3-Embedding-8B     & 9.32      & 9.32   & 9.32  & 0.30 \\ \hline
\multicolumn{1}{c|}{\multirow{4}{*}{\makecell{Heuristic \\ Metadata Augmentation}}}   & Session Clustering & 4.84 & 4.84 & 4.84 & 0.60 \\
\multicolumn{1}{c|}{} & User+Session Clustering & 8.94            & 8.94            & 8.94           & 1.05             \\
\multicolumn{1}{c|}{}     & Time Proximity         & 6.67      & 6.67   & 6.67  & 0.45 \\
\multicolumn{1}{c|}{}     & User+Spatiotemporal         & 8.80      & 8.80   & 8.80  & 0.75 \\ \hline

\multicolumn{1}{c|}{\multirow{5}{*}{\makecell{VLM Two-Stage \\ Decompose \& Rerank}}} & Qwen2.5-VL-72B          & 11.39           & 10.16           & 10.67          & 0.15             \\
\multicolumn{1}{c|}{}     & Qwen3-VL-235B          & 15.78     & 14.35  & 14.95 & 0.90 \\
\multicolumn{1}{c|}{}     & Gemini-3-Flash & 20.89     & 15.53  & 17.06 & 0.60 \\
\multicolumn{1}{c|}{}     & GPT-4o                 & 18.50     & 17.34  & 17.84 & 0.60 \\
\multicolumn{1}{c|}{}     & Claude-Sonnet-4.5      & \underline{24.79} & \underline{23.01}  & \underline{23.74} & \underline{1.50} \\ \hline
\multicolumn{2}{c|}{\model\ (Ours)}                                                                         & \textbf{30.95}  & \textbf{30.46}  & \textbf{30.28} & \textbf{7.20}    \\
\multicolumn{2}{c|}{Relative Improvement}                                                         & 24.8\% & 32.4\% & 27.5\% & 380.0\% \\ \hline
\bottomrule
\end{tabular}}

\caption{Performance comparison of different methods on \dataset. Best result is given in bold, and the second best is underlined. \textit{Relative Improvement} is computed against the best baseline result.
% ``\textit{Closed}" means the model is close-sourced.
}

\label{tab:main_table}

\vspace{-7pt}
\end{table*}

\section{Experiments}

\subsection{Experimental Setup}

\paragraph{Evaluation Metrics.}
% Unlike traditional image retrieval tasks that evaluate a fixed-length ranked list, Bundle Image Retrieval (IBC) requires the model to return a compact, cohesive set of images. Since the optimal bundle size varies across queries and the length of the predicted bundle dynamically depends on the model's output, we evaluate performance using set-level metrics. 
% Specifically, we use \textbf{Precision}, \textbf{Recall}, and \textbf{F1 score}. Additionally, we report \textbf{Exact Match (EM)}, which measures the percentage of queries where the predicted bundle is perfectly identical to the ground-truth bundle.

Unlike traditional ranked lists, IBC outputs dynamic-length cohesive image sets. Thus, we evaluate performance using set-level \textit{Precision}, \textit{Recall}, and \textit{F1 score}, alongside \textit{Exact Match (EM)}, \ie, the percentage of perfectly predicted ground-truth bundles. The metrics are evaluated at per-query-level and reported with the average.

\paragraph{Baselines.}
Since IBC is a novel task, we establish comprehensive baselines of four different types:
\begin{itemize}[leftmargin=10pt]
    \item \textbf{Multimodal Embedding:} Directly retrieves images using state-of-the-art vision-language multimodal embedding models, including CLIP-ViT-B/32~\cite{radford2021learning}, SigLIP2-giant~\cite{tschannen2025siglip}, Qwen3-VL-Embedding-8B~\cite{li2026qwen3}, and RzenEmbed-7B~\cite{jian2025rzenembed}.
    \item \textbf{Caption + Text Embedding:} First generates captions for all images using GPT-4o, then performs text-to-text retrieval using BM25~\cite{robertson2025bm25}, BGE-M3~\cite{chen2024bge}, and Qwen3-Embedding-8B~\cite{zhang2025qwen3}.
    \item \textbf{Heuristic Metadata Augmentation:} Applies heuristic re-ranking on multimodal embedding results using  metadata, \ie, session clustering, user+session clustering  time proximity boosting and user+spatiotemporal.
    \item \textbf{VLM Two-Stage Decompose \& Rerank:} A strong agentic baseline where a VLM first decomposes the query into multiple sub-queries, retrieves top-$K$ candidates for each independently, and then reranks the final bundle in the candidate pool. Evaluated VLMs include GPT-4o~\cite{hurst2024gpt}, Claude-Sonnet-4.5-20250929~\cite{anthropic2025claudesonnet45}, Gemini-3-Flash~\cite{googledeepmind2025gemini3flash}, Qwen2.5-VL-72B~\cite{bai2025qwen25vltechnicalreport} and Qwen3-VL-235B~\cite{bai2025qwen3}.
\end{itemize}
For baselines that naturally return a ranked list of individual images (\eg, multimodal/text embeddings and metadata augmentation), we follow previous works~\cite{xu2026photobench, deng2026deepimagesearch} and adopt an \textit{oracle-size evaluation}: for each query, we truncate the retrieved list to the top-$|B^\ast|$ images, where $|B^\ast|$ denotes the cardinality of the GT bundle. Consequently, since the predicted and GT sets have identical sizes, their Precision, Recall, and F1 scores become mathematically equivalent.

\paragraph{Implementation Details.} Due to the page limitation, we move the implementation details to Appendix~\ref{appendix: implementation details}.

\subsection{Main Results}

Table~\ref{tab:main_table} summarizes the performance of \model\ and baselines. We make the following observations:
\begin{itemize}[leftmargin=10pt]
    \item \textbf{Traditional point-wise matching is fundamentally inadequate for IBC.} Even with model scaling, their performance remains low. This indicates that IBC cannot be solved by simply measuring independent text-image alignment, and cross-image relations are completely lost in atomic point-wise matching paradigms.
    
\item \textbf{Naive metadata augmentation is insufficient.} Although spatiotemporal locality naturally correlates with real-world events, explicitly applying it as a rigid heuristic filter into baseline models actually degrades their  F1. This degradation highlights that spatiotemporal proximity alone cannot fulfill the specific relational roles demanded by the query, reinforcing the indispensable role of active inline reasoning in IBC.
    
    \item \textbf{Static decomposition lacks relational constraints.} The decompose-and-rerank paradigm  yields a  boost in  Recall and Precision, as decomposing the query ensures a broader coverage of the visual narrative. However, its EM rate remains devastatingly low. Since sub-queries are retrieved \textit{independently}, the system cannot enforce vital inter-image constraints, highlighting the cross-image composition nature of IBC.
    
    \item \textbf{\model\ excels via incremental composition.} \model\ achieves state-of-the-art performance across all metrics. This comprehensive improvement is attributed to its dynamic hyperedge discovery: by adaptively expanding hyperedges via parallel beam search, \model\ balances step-wise precision with whole-bundle coherence, effectively enforces the cross-image relational constraints that other paradigms ignore.
\end{itemize}

\subsection{Component Ablation}

\begin{table}[t]

\centering

% \vspace{-11pt}
% \footnotesize
\resizebox{0.49\textwidth}{!}{

\begin{tabular}{c|cccc}
\toprule
\hline
Method          & Precision & Recall & F1    & EM   \\ \hline
GPT-4o (baseline) & 18.50 & 17.34 & 17.84 & 0.60 \\ \hline
\model\ (Ours) & \textbf{30.95} & \textbf{30.46} & \textbf{30.28} & \textbf{7.20} \\
w/o Diverse Seed                      & \underline{27.52}    & \underline{27.66}    & \underline{27.21}    & \underline{6.45}    \\
w/o Candidate Pruning & 24.57 & 24.74 &  24.69 & 5.85 \\
w/o Beam Search & 25.63     & 25.93  & 25.35 & 6.30 \\
w/o VLM Rerank  & 26.33     & 26.70  & 26.13 & 6.15 \\
\hline
\bottomrule
\end{tabular}

}

\caption{Ablation study of different components of our proposed \model.
}
\label{tab:component ablation}

\vspace{-7pt}

\end{table}

In Table~\ref{tab:component ablation}, we  ablate the three core stages of \model\  to isolate their contributions. We can observe that:
\begin{itemize}[leftmargin=10pt]
    \item Relying on a single starting point artificially restricts the search scope. The performance degradation without diverse seeding can likely be attributed to the search getting trapped in local optima, whereas seeding the search from visually and semantically diverse anchors helps ensure broader coverage across the global image pool.

    \item Omitting Contextual Candidate Pruning leads to a clear performance drop. Without this constraint, each expansion step must search over a much larger and noisier candidate space, making the beam search more likely to select visually plausible but relationally incompatible distractors. Importantly, the resulting model still outperforms the vanilla GPT-4o baseline by a large margin, suggesting that the performance of \model\ cannot be attributed to metadata pruning alone. Instead, contextual pruning provides a tractable candidate neighborhood, while adaptive expansion and whole-bundle verification remain essential for composing relationally coherent bundles.

    \item Downgrading the parallel beam search to a greedy expansion strategy drastically hurts performance. Because greedy choices in a massive combinatorial space are highly susceptible to early search errors, maintaining a diverse beam of hypothesis paths appears essential for uncovering the promising visual story.

    \item Removing the VLM pointwise reranking and relying purely on embedding-based path scores causes performance drop. This suggests that while dense embeddings are efficient for guiding the search direction, they cannot reliably verify complex structural logic or instance-level identity consistency across the completed bundle.

\end{itemize}

\begin{figure}[t]
% \vspace{-3pt}
    \centering
    \includegraphics[width=0.99\linewidth]{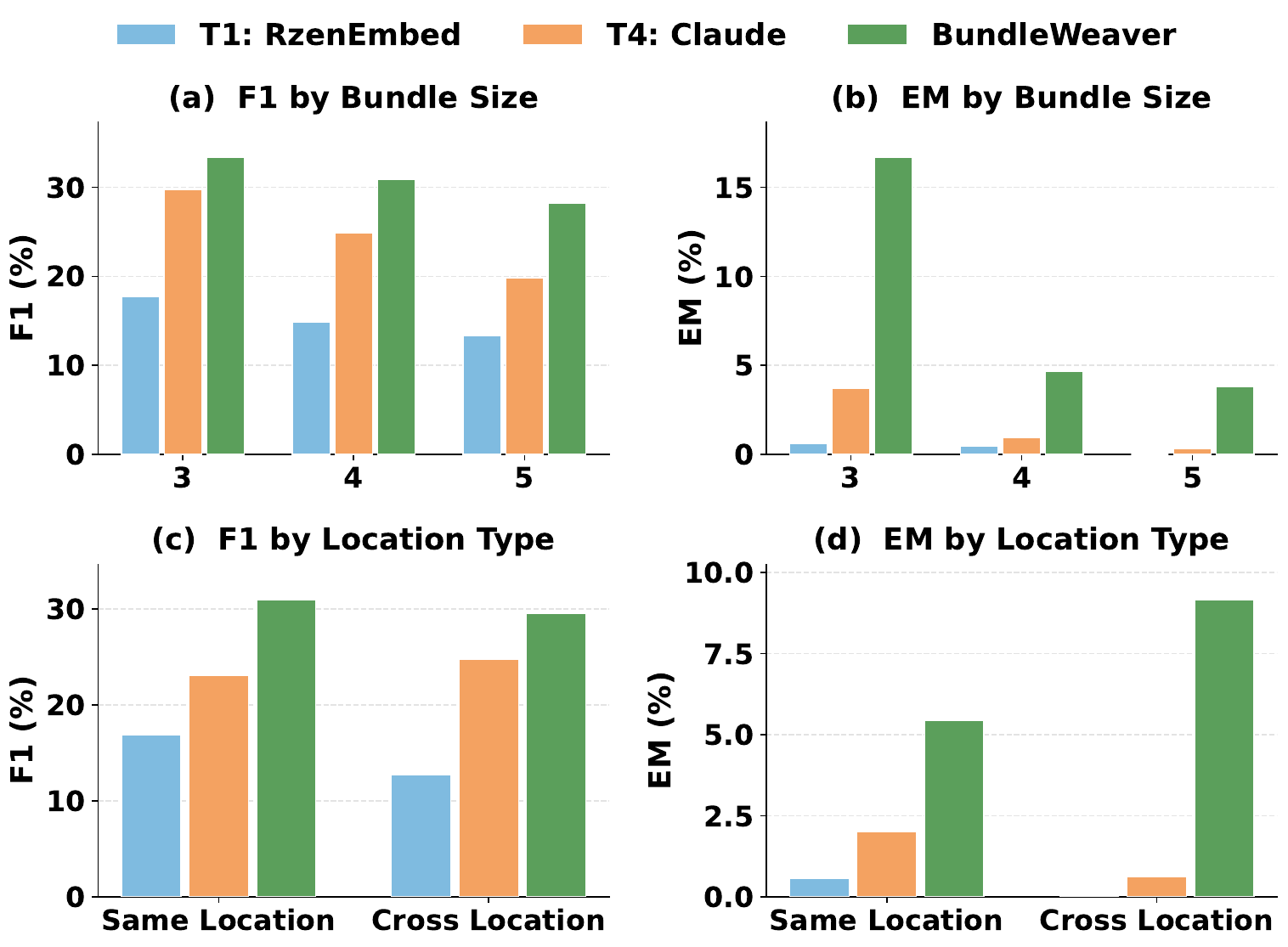}
    \caption{Breakdown analysis on bundle size and location type.}
    \label{fig: breakdown}

    % \vspace{-3pt}
\end{figure}

\subsection{Breakdown Analysis}

Figure~\ref{fig: breakdown} further breaks down performance by bundle size and location type. Across all bundle sizes, \model\ consistently outperforms the baselines. As the bundle size increases, all methods become worse, indicating that IBC becomes harder when more images must be jointly recovered. However, \model\ shows a smaller performance drop, suggesting that incremental bundle expansion is more robust than independently retrieving images for decomposed sub-queries.

 RzenEmbed performs notably worse on cross-location bundles, since visually diverse images are difficult to connect through point-wise similarity alone. The VLM decompose-and-rerank baseline improves F1, but its EM remains low, showing that static decomposition can retrieve partially relevant images but often fails to assemble the exact coherent bundle. In contrast, \model\ achieves the best metrics for both location types. These results indicate that explicitly modeling missing relational roles and verifying the whole bundle are important for robust bundle retrieval.

\subsection{Backbone Generalizability}

% \usepackage{multirow}

\begin{table}[t]

\centering

% \vspace{-5pt}
% \footnotesize
\resizebox{0.49\textwidth}{!}{

\begin{tabular}{c|c|cccc}
\toprule
\hline
Backbone & Method & Precision & Recall & F1 & EM \\ \hline
\multirow{2}{*}{Qwen3-VL-235B} & Dec. \& Re. & 15.78 & 14.35 & 14.95 & 0.90 \\
 & \model & 25.99 & 26.23 & 25.73 & 5.25 \\ \hline
\multirow{2}{*}{Gemini-3-Flash} & Dec. \& Re. & 20.89 & 15.53 & 17.06 & 0.60 \\
 & \model & 25.52 & 26.52 & 25.55 & 4.35 \\ \hline
\multirow{2}{*}{Claude Sonnet 4.5} & Dec. \& Re. & 24.79 & 23.01 & 23.74 & 1.50 \\
 & \model & \textbf{31.12} & \underline{30.39} & \textbf{30.32} & \underline{6.60} \\ \hline
\multirow{2}{*}{GPT-4o} & Dec. \& Re. & 18.50 & 17.34 & 17.84 & 0.60 \\
 & \model & \underline{30.95} & \textbf{30.46} & \underline{30.28} & \textbf{7.20} \\ \hline
\bottomrule
 
\end{tabular}

}
\caption{Performance w.r.t. different VLM backbones. \textit{Dec. \& Re.} represents the baseline \textit{VLM Two-Stage Decompose \& Rerank}.
}
\label{tab:backbone}

\vspace{-3pt}
\end{table}

To verify the generalizability of our framework, we evaluate \model\ across various frontier VLM backbones. As shown in Table~\ref{tab:backbone}, \model\ consistently outperforms the agentic decompose-and-rerank baseline across diverse architectures, from open-weight models to close-source giants. Notably, the EM rate experiences a massive surge regardless of the chosen backbone. This universal leap confirms that the relational blindness inherent in independent decomposition cannot be rescued merely by deploying a smarter VLM. Instead, it is the structural design of \model, \ie, incremental hyperedge discovery coupled with whole-bundle verification, that effectively enforces bundle-level constraints.

\subsection{More Experiments}

We provide more experiments in Appendix~\ref{appendix: more experiments}.
We further address potential questions about our paper in Appendix~\ref{appendix: response}.
\section{Related Works}
\label{sec:related_works}

% Traditional T2I retrieval formulates cross-modal matching as an independent, point-wise scoring task, established by benchmarks like MSCOCO and Flickr30K~\cite{lin2014microsoft, plummer2015flickr30k}. Subsequent benchmarks have introduced richer contexts, including composed image retrieval~\cite{wu2021fashion, baldrati2023zero}, lifelog retrieval~\cite{gurrin2023introduction}, and multi-hop visual-history search~\cite{xu2026photobench, deng2026deepimagesearch}. Alongside benchmark evolution, retrieval architectures have advanced from foundational dual-encoders (\eg, CLIP, SigLIP)~\cite{radford2021learning, tschannen2025siglip} to powerful MLLM-based embeddings (\eg, Qwen3-VL, RzenEmbed)~\cite{li2026qwen3, jian2025rzenembed}. Recently, agentic retrieval methods have further replaced one-shot embedding search with reasoning-driven iterative query decomposition~\cite{jiang2024mmsearch, wang2026oscar, liang2026mm}. Despite these advancements, existing retrieval paradigm remains atomic. IBC redefines the retrieval objective, shifting from independently scoring images to composing cohesive image bundles.

\subsection{Vision-Language Retrieval}

Vision-language retrieval has traditionally been formulated as a cross-modal matching problem, where the relevance between a query and each candidate image is independently estimated. Early benchmarks such as MSCOCO and Flickr30K established this point-wise retrieval paradigm by evaluating text-to-image and image-to-text alignment~\cite{lin2014microsoft, plummer2015flickr30k}. Recent benchmarks have expanded retrieval scenarios beyond simple image-text matching, introducing additional contexts such as composed image retrieval~\cite{wu2021fashion, baldrati2023zero}, lifelog retrieval~\cite{gurrin2023introduction}, and multi-hop visual-history search~\cite{xu2026photobench, deng2026deepimagesearch}. These tasks improve retrieval complexity by incorporating references, personal histories, or contextual constraints~\cite{lin2025can}. However, their retrieval targets remain individual images or moments whose relevance can be evaluated independently.

Alongside benchmark development, retrieval models have evolved from dual-encoder architectures, such as CLIP and SigLIP~\cite{radford2021learning, tschannen2025siglip}, to more expressive multimodal large language model (MLLM)-based representations, such as Qwen3-VL and RzenEmbed~\cite{li2026qwen3, jian2025rzenembed}. More recently, agentic retrieval approaches have introduced iterative reasoning, query decomposition, and verification to improve retrieval under complex queries~\cite{jiang2024mmsearch, wang2026oscar, liang2026mm}. Despite these advances, existing retrieval systems largely follow an atomic retrieval paradigm: images are ranked independently based on their individual relevance to the query.

In contrast, IBC introduces a new retrieval objective by shifting the target from isolated image relevance to cohesive image bundle composition. The validity of an IBC result depends not only on whether each image matches the query, but also on whether the retrieved images jointly satisfy relational, temporal, and narrative constraints. This requires reasoning over interactions among images rather than independent cross-modal matching.

\subsection{Beyond Atomic Image Retrieval}

Several related tasks have explored retrieving multiple images or reasoning over visual collections. Multi-image retrieval extends traditional text-to-image retrieval by returning multiple relevant images, while lifelog retrieval focuses on finding relevant moments from personal photo streams~\cite{gurrin2023introduction}. Composed image retrieval further incorporates additional visual references to refine the retrieval target~\cite{wu2021fashion, baldrati2023zero}. However, these tasks still evaluate retrieval results primarily at the image level, where each retrieved item contributes independently to the final ranking.

Visual storytelling~\cite{huang2016visual, hu2020makes} represents another related direction, but it differs from IBC in its objective. Visual storytelling assumes that the input images are already provided and focuses on generating coherent textual narratives. In contrast, IBC starts from a text query and requires the system to discover and assemble the image set itself.

Therefore, while existing retrieval and generation tasks have progressively incorporated richer contexts, none explicitly address the problem of retrieving a relationally coherent image bundle from a large unstructured image pool. IBC fills this gap by treating bundle-level consistency as the fundamental retrieval criterion.
\section{Conclusion}
In this paper, we introduce \textbf{Image Bundle Composition (IBC)}, which shifts the retrieval objective from ranking isolated snapshots to dynamically composing cohesive bundles bound by explicit cross-image relations. To establish this new paradigm and mitigate the computational bottlenecks of combinatorial annotation, we build \dataset, the first meticulously verified benchmark dataset for this task. Furthermore, we propose \model, an agentic framework that reformulates IBC as query-conditioned incremental hyperedge discovery. By leveraging adaptive sub-query generation, parallel beam search, and whole-bundle VLM reranking, \model\ effectively enforces cross-image structural constraints and significantly outperforms existing baseline paradigms.

\section*{Limitations}

As a pioneering effort in formulating the Image Bundle Composition (IBC) paradigm, this work naturally possesses certain limitations that offer exciting avenues for future research. First, our current investigation is primarily constrained to static images within personal photo collections (\ie, YFCC). We have not yet extended this paradigm to more specialized domains (\eg, medical imaging sequences or legal evidentiary archives) or dynamic modalities such as Video Bundle Retrieval, where temporal dynamics are continuous rather than discrete. 
Second, \model\ currently operates in a training-free, zero-shot manner by leveraging off-the-shelf foundation models. While this elegantly demonstrates the generalized relational reasoning power of our incremental search mechanism, we have not yet explored end-to-end fine-tuning strategies on specific datasets. Developing parameter-efficient tuning methods tailored specifically for IBC remains a highly promising direction for future work.

\section*{Acknowledgments}
This paper is supported by National Natural Science Foundation of China (624B2096, 72595872, 72542012, 62322603).

\bibliography{custom}

\appendix

\newpage

\section{Ethical Considerations}

The foundation of our benchmark, \dataset, relies on the public YFCC-100M corpus~\cite{thomee2016yfcc100m}. To ensure strict adherence to data privacy and copyright standards, we exclusively utilize images distributed under Creative Commons licenses that explicitly authorize research applications. We have also audited the associated spatiotemporal metadata to ensure compliance. The release of our dataset will rigorously follow all original redistribution guidelines to respect and protect content creators' rights.

As Image Bundle Composition (IBC) introduces the capability to model complex spatiotemporal trajectories and bundle-level relational constraints, it is crucial to address potential privacy implications. Within the scope of this dataset, privacy risks are minimized because the visual assets are already publicly available under Creative Commons licenses, and our cross-image modeling focuses purely on advancing multimodal relational reasoning rather than profiling real-world identities. More importantly, the core motivation of the IBC paradigm is user-centric. The technology is envisioned as an intelligent agent for personal photo management, empowering individuals to dynamically compose visual stories within their \textit{own} private galleries or encrypted local storage. It is neither intended for, nor optimized for, the unauthorized surveillance or analysis of external, third-party subjects.

\section{Impact Discussion}
While \model\ achieves remarkable relative improvement in Exact Match (EM) over the strongest baseline, we  acknowledge that the absolute performance remains modest. This substantial headroom underscores the intrinsic difficulty and profound significance of the IBC task. Perfect recovery of a non-decomposable visual narrative from a massive $\mathcal{O}(N^K)$ combinatorial space poses a formidable challenge for current multimodal agents, exposing a critical blind spot in contemporary retrieval methodologies. Consequently, \model\ is positioned not as a definitive endpoint, but as a foundational baseline and a pioneering starting point. We hope our framework and the  benchmark will attract broader attention from the research community to tackle this challenging yet highly meaningful problem. Ultimately, IBC highlights a profound paradigm shift for the community, demonstrating that future retrieval systems must evolve from passive, point-wise rankers into active, reasoning-driven constructors capable of inline relational composition.

\section{More Details of Dataset Construction}
\label{appendix: dataset construction}

\begin{figure*}[t]
    \centering
    \vspace{-7pt}
    \includegraphics[width=0.99\linewidth]{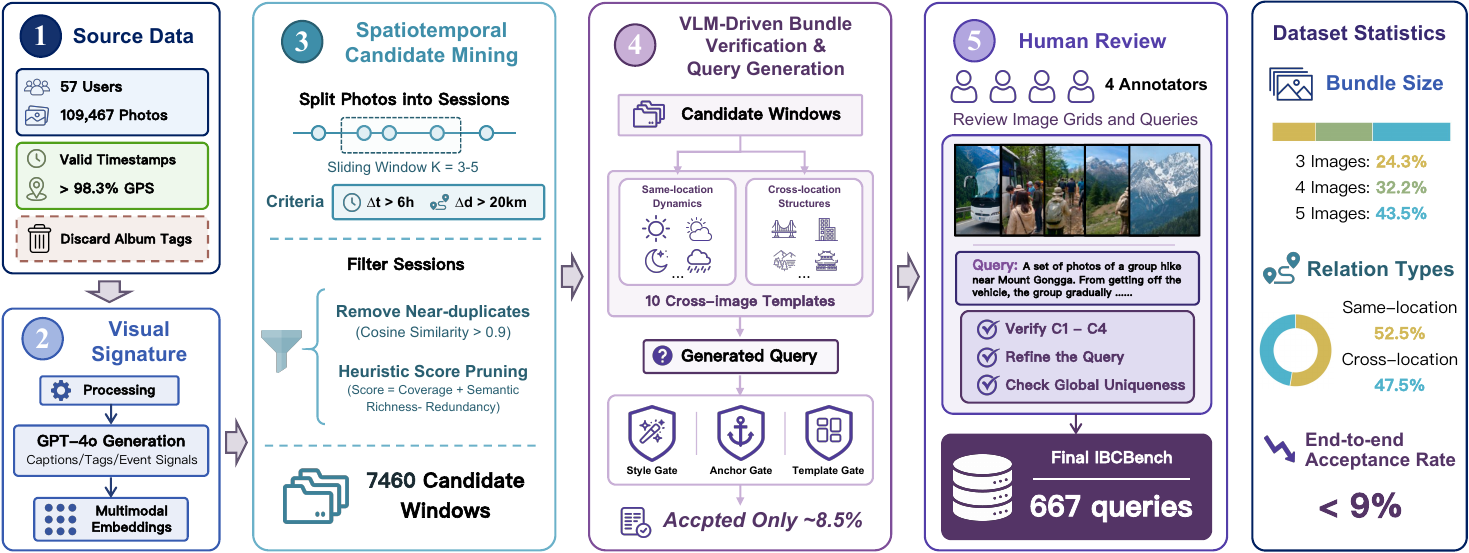}
    % \vspace{-3pt}
    \caption{The demonstration of dataset construction pipeline for \dataset.}
    \vspace{-5pt}
    \label{fig: dataset construction}
\end{figure*}

\subsection{Property Requirements of a Valid Image Bundle}
\label{sec: bundle requirements}

Before mining candidates, we formally define the criteria for a high-quality bundle. A candidate subset $B \subseteq \mathcal{I}$ must satisfy four strict criteria to be accepted:
\begin{itemize}[leftmargin=10pt]
    \item \textbf{(C1) Joint Completeness:} Every image in $B$ must play an indispensable role in fulfilling the query. Removing any single image would break the narrative or structural integrity of the answer.
    \item \textbf{(C2) Cross-image Binding:} The query $q$ must enforce a higher-order relation across the images (\eg, temporal progression, recurring rituals, or before-and-after contrast), rather than merely listing independent visual targets.
    \item \textbf{(C3) Uniqueness within Pool:} To ensure robust evaluation, $B$ must be  unambiguous. There must be minimized subset in the entire image pool $\mathcal{I}$ that satisfies $q$ better than or equally well as $B$.
    \item \textbf{(C4) Bounded Redundancy:} The images within $B$ must be visually distinct and not near-duplicates (\eg, burst shots).
\end{itemize}

\subsection{Source Data and Visual Profiles}
We source our raw image pool from the YFCC-100M dataset~\cite{thomee2016yfcc100m}, selecting 57 users with rich metadata. The global pool comprises 109,467 photos, all possessing valid EXIF timestamps and over 98.3\% having GPS coordinates. Crucially, we discard all user-defined album ids from YFCC, ensuring that our bundles emerge from visual and spatiotemporal relations rather than pre-existing manual groupings.

To enable semantic filtering, we cache rich visual signals for all 109,467 images. We utilize GPT-4o~\cite{hurst2024gpt} to extract dense text captions, word-level tags, and event-level signals (\eg, \textit{wedding, summit}), whose prompt is shown in Appendix~\ref{appendix: visual profile prompt}. Additionally, we compute multimodal embeddings using RzenEmbed, which serves to penalize redundancy during the candidate mining phase.

\subsection{Spatiotemporal Candidate Mining}
\label{sec: candidate mining}

To bypass the $\mathcal{O}(N^K)$ combinatorial explosion, we anchor our search space using metadata constraints. We first group each user's photos into spatiotemporal sessions, breaking sequences wherever there is a time gap $\Delta t > 6$ hours or a geographic jump $\Delta d > 20$ km. 

Within each session, we enumerate sliding windows of size $K \in [3, 5]$. To prioritize structurally promising candidates, we evaluate each window using a composite heuristic score:
\begin{equation}
\label{eq: heuristic score}
\text{Score}
=S_{\text{st}} + S_{\text{sem}} - P_{\text{red}},
\end{equation}
where \(S_{\text{st}}\), \(S_{\text{sem}}\), and \(P_{\text{red}}\) denote spatiotemporal coverage, semantic richness, and redundancy penalty, respectively. The three components are detailed below:
\begin{itemize}[leftmargin=10pt]
    \item \textbf{Spatiotemporal Coverage ($S_{\text{st}}$).} A valid visual story typically requires a meaningful progression in time or space, rather than a static burst of photos taken simultaneously. The spatiotemporal coverage component rewards candidate windows that span a reasonable duration and physical distance:
\begin{equation}
\begin{aligned}
S_{\text{st}}(W)
={}& \min\left(2, \frac{T}{2}\right)
+ \min\left(2, \frac{D}{3}\right) \\
&+ \mathbf{1}_{\text{anchor}}(W).
\end{aligned}
\end{equation}
where $T$ denotes the total time span of the window in hours, and $D$ represents the cumulative geodesic distance in kilometers between consecutive images. The indicator function $\mathbf{1}_{\text{anchor}}(W)$ returns 1 if the photos share at least one identical reverse-geocoded address prefix, which rewards spatial anchoring. The $\min(\cdot)$ operations act as clipping functions to prevent extreme outliers (\eg, intercontinental flights) from disproportionately dominating the score.

\item \textbf{Semantic Richness ($S_{\text{sem}}$).}
Beyond physical progression, a high-quality bundle must encapsulate a rich set of visual and narrative elements. We leverage the pre-cached tags and event signals generated by GPT-4o to measure semantic diversity:
\begin{equation}
    S_{\text{sem}}(W) = \min\left(2, \frac{|\mathcal{T}_W|}{6}\right) + \min\left(1.5, \frac{|\mathcal{E}_W|}{4}\right)
\end{equation}
where $|\mathcal{T}_W|$ and $|\mathcal{E}_W|$ denote the cardinality of the union of all word-level tags and event-level signals within the window $W$, respectively. This component explicitly favors bundles that exhibit diverse semantic concepts, ensuring that the retrieved images are information-dense.

\end{itemize}

Candidate windows with a total score below a strict threshold (default to 2.5 in our implementation) are discarded. This heuristic pruning aggressively reduces the search space, yielding a manageable pool of 7,460 highly diverse candidate windows.

\subsection{VLM-Driven Bundle Verification and Query Generation}
The heuristically mined windows satisfy spatiotemporal constraints, but they do not necessarily exhibit narrative coherence (C1-C2). We employ Claude-Opus-4.5~\cite{anthropic2025claudeopus45} as a rigorous semantic verifier. The VLM is instructed to evaluate each candidate against several predefined cross-image templates divided into two families:
\begin{itemize}[leftmargin=10pt]
    \item \textbf{Same-location Dynamics:} \eg, visible state changes, event phase progressions (preparation $\rightarrow$ climax $\rightarrow$ end), or  physical transitions (day-to-night).
    \item \textbf{Cross-location Structures:} \eg, repeating travel rituals, route-level continuity, or shared thematic activities across different venues.
\end{itemize}
For a candidate to be accepted, the VLM must successfully map it to one of these templates, generate a natural language query, and articulate a specific, instance-level \textit{shared anchor} (required by C2, \ie, \textit{the same kayaker}). To prevent the VLM from generating lazy enumerations, we implement a three-layer programmatic gate (\ie, Style Gate, Anchor Gate, Template Gate, see Appendix~\ref{appendix: program gate} for the prompt) that automatically rejects outputs containing forbidden grammatical structures (\eg, strings of isolated gerunds) or vague anchors. This strict verification accepts only about 8.5\% of the candidates.

\subsection{Human Review}
In the final stage, all VLM-approved candidates undergo rigorous human review. Four expert annotators examine the image grid, the generated query, and the global pool context. Annotators are tasked with verifying constraints C1-C4, refining the query text for naturalness, and explicitly rejecting any queries that lack pool-wide uniqueness (C3).

\begin{figure}[t]
    \centering

    \begin{subfigure}{0.95\linewidth}
        \centering
        \includegraphics[width=\linewidth]{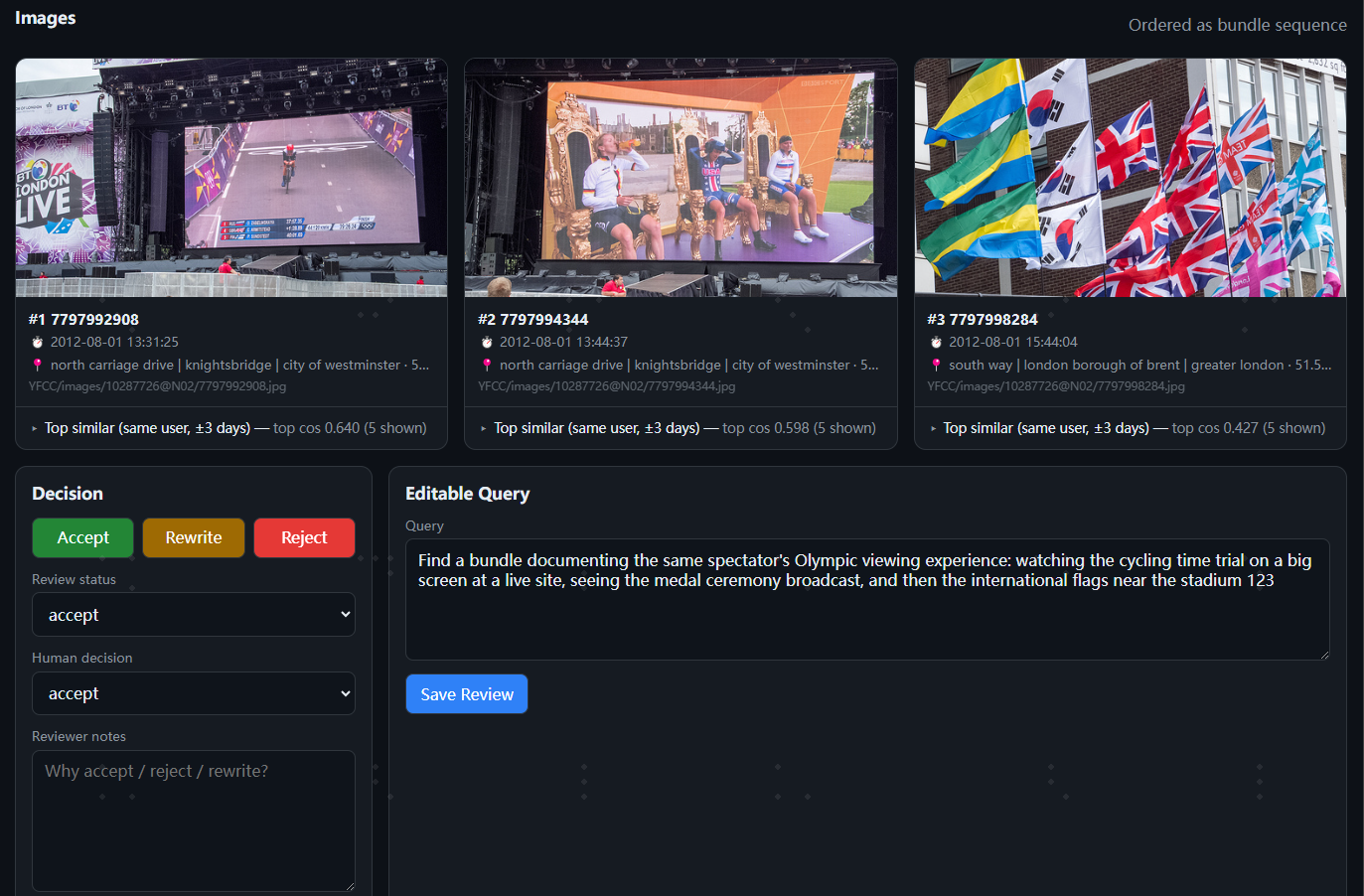}
        \label{fig:app_labelling_web}
    \end{subfigure}

    \vspace{-5pt}

    \begin{subfigure}{0.95\linewidth}
        \centering
        \includegraphics[width=\linewidth]{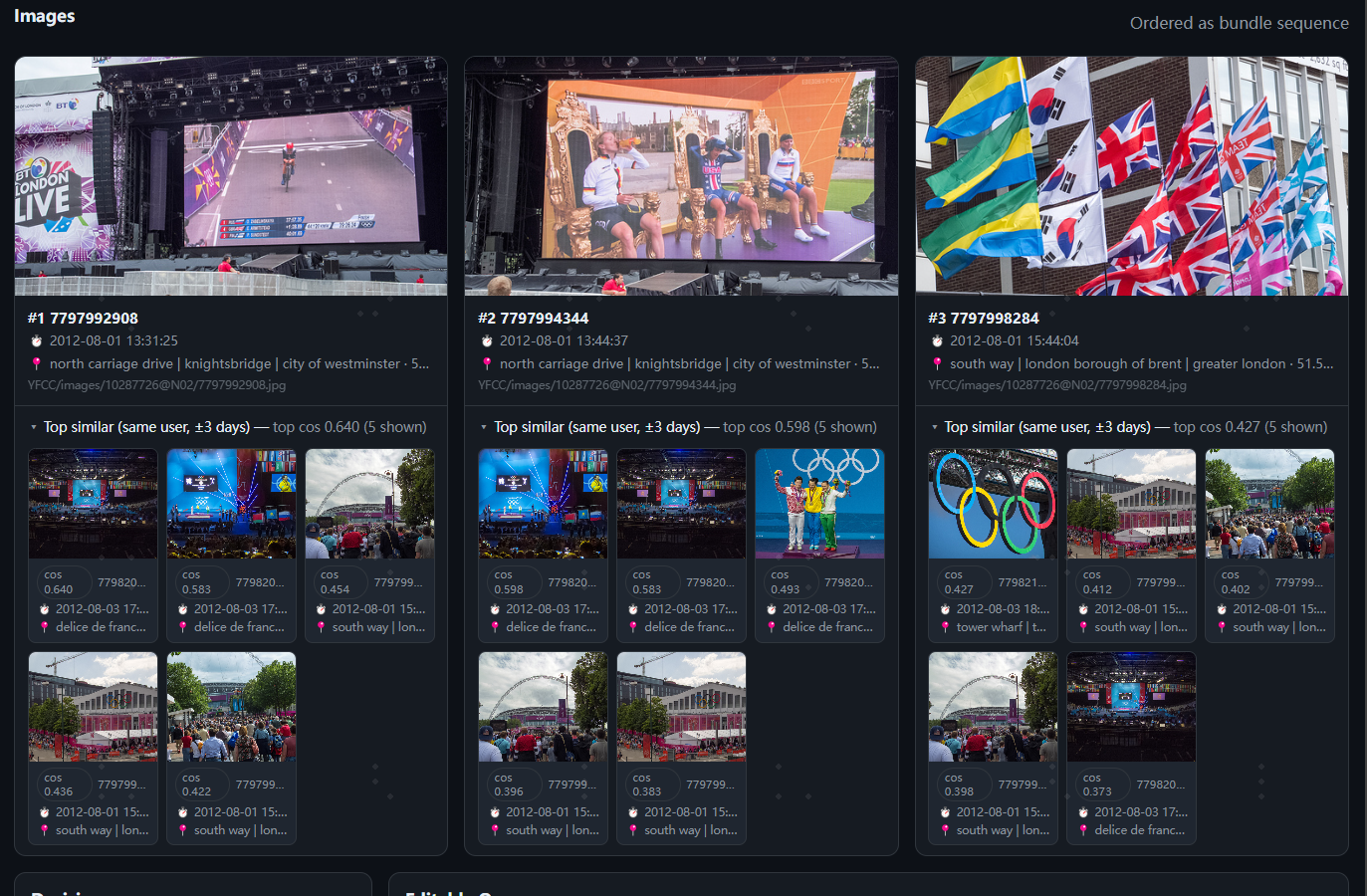}
        \label{fig:app_labelling_dedup}
    \end{subfigure}
    \vspace{-5pt}

    \caption{The web demo for human review and deduplication.}
    \label{fig: review web}
\end{figure}

Specifically, for each candidate, the interface (shown in Figure~\ref{fig: review web}) displays:
(i) the generated query and VLM rationale (including \texttt{template\_id}, \texttt{shared\_anchor}, and all audit fields); 
(ii) the bundle images rendered as a grid with per-image metadata (capture time, GPS-derived address, and device);
and (iii) a \emph{uniqueness panel} that supports the annotator in verifying constraint (C3).

\paragraph{Embedding-based uniqueness assistance.}
A key challenge for human reviewers is judging whether a given bundle is the \emph{unique best answer} within a pool of 109K images, since visually similar alternatives may exist.
To assist this decision, we precompute pairwise cosine similarities using the Rzenembed embedding model and, for each bundle image, retrieve its top-5 most similar photographs from the \emph{same user} within a $\pm$3-day time window.
These nearest neighbors are displayed alongside the bundle image with their similarity scores and metadata, enabling reviewers to quickly assess whether a near-duplicate or semantically interchangeable image exists that would undermine the bundle's uniqueness.
A global \emph{uniqueness risk} indicator is also computed as the maximum neighbor similarity across all bundle images, color-coded as \textsc{high} ($\cos \geq 0.85$), \textsc{medium} ($0.75 \leq \cos < 0.85$), or \textsc{low} ($\cos < 0.75$).

\paragraph{Annotator actions.}
For each candidate, the reviewer selects one of three decisions:
\textsc{Accept} (the bundle satisfies all four constraints and the query is accurate),
\textsc{Reject} (any constraint is violated), or
\textsc{Rewrite} (the bundle is valid but the query needs editing).
When accepting or rewriting, the annotator may edit the query text to improve naturalness or specificity.
Free-text notes are also supported for flagging borderline cases.

\paragraph{Agreement and yield.}
A total of 1,674 candidates were reviewed, of which 667 (39.8\%) were accepted.
Per-annotator acceptance rates range from 30.2\% to 49.9\%, reflecting differences in individual strictness during the initial review.
To reduce annotator-specific bias, accepted candidates were further cross-validated by other annotators, and borderline or disputed cases were discussed collectively before inclusion in the final dataset.
Moreover, we leverage the upstream programmatic gates and VLM audit fields to enforce a consistent quality floor throughout the review process.

Furthermore, we manually inspect the predicted bundles from \model\ and the baseline that were automatically evaluated as incorrect. As detailed in Section~\ref{appendix: human validation}, the rate of genuine false negatives (\ie, unannotated but reasonable alternative bundles) is negligible. This further suggests the uniqueness of the ground-truth bundles established during the dataset construction process.

\subsection{Final Dataset Statistics}

Following human curation, the final \dataset\  dataset contains \textbf{667 high-quality queries} evaluated against the massive \textbf{109,467-image} pool. The bundle sizes are distributed across 3 images (24.3\%), 4 images (32.2\%), and 5 images (43.5\%). The semantic relations are well-balanced, with roughly 52.5\% focusing on same-location dynamics  and 47.5\% capturing cross-location structural ties. Through this exhaustive pipeline with an end-to-end acceptance rate under 9\%, we aim to ensure that every ground-truth bundle in the dataset is unique and cohesive that challenges the limits of modern retrieval systems.

\tcbset{
    promptstyle/.style={
        colback=backorange,
        colframe=frameorange,
        fonttitle=\bfseries,
        arc=2pt,
        breakable, % 允许跨页断行
        left=2mm, right=2mm, top=2mm, bottom=2mm,
        boxrule=0.5pt,
        fontupper=\small,
        fontlower=\small
    }
}

\subsection{Prompt Demonstration}

\subsubsection{Prompt for Building Visual Profiles}
\label{appendix: visual profile prompt}

\begin{tcolorbox}[promptstyle, title=Image Profiling]
\textbf{System:} You are helping construct Image Bundle Composition data. Return strict JSON with keys: caption, tags, event\_signals, bundle\_notes. Keep everything in English.

\textbf{User:} Describe this image for downstream bundle mining. Match this JSON schema exactly:

\{
"caption": "Detailed but concise English description of the image.", "tags": ["short tags capturing objects, scene and activity"], "event\_signals": ["possible event or state-change clues"], "bundle\_notes": "Short note about whether the image may help a time/state/route/event progression bundle."\}
\end{tcolorbox}

\subsubsection{Prompt for Bundle Verification and Query Generation}
\label{appendix: verify prompt}

\begin{tcolorbox}[promptstyle, title=Bundle Verification and Query Generation]
\textbf{System:} You are a strict IBC dataset judge. Follow every rule literally.

\textbf{User:} You are a STRICT validator for a Image Bundle Composition (IBC) dataset.
The bundle is ACCEPTED only if ALL of the following hold:
\begin{itemize}[leftmargin=10pt]
    \item[1.] The images fit exactly one template in the allowed list below.
    \item[2.] There is a CONCRETE instance-level shared\_anchor (specific person / object / ritual /
     named route / repeated action visible across images). Generic phrases like 'temporal progression',
     'same day', 'the trip', 'the route', 'cross-image reasoning' are NOT acceptable anchors.
     \item[3.] Removing any single image would clearly break the bundle's meaning.
     \item[4.] The query describes one UNIQUE bundle within the available pool. It is not an album dump.
     \item[5.] The query does NOT read like a list of independent single-image targets.
\end{itemize}

FORBIDDEN QUERY PATTERNS (auto-reject):
\begin{itemize}[leftmargin=10pt]
    \item 'showing A, B, C, and D' style enumeration where each item is a separate scene
    \item 'from A, through B, to C, and finally D' when A/B/C/D are heterogeneous scenes without a shared anchor
    \item 4+ gerund verbs in a comma/and list ('doing X, doing Y, doing Z, doing W')
\end{itemize}

SAME-LOCATION FAMILY RULES:
\begin{itemize}[leftmargin=10pt]
    \item The defining logic must come from what CHANGES, PROGRESSES, or becomes STRUCTURALLY
  DIFFERENTIATED within one location, not from the fact images share a place.
  \item Reject queries that are a list of different objects/views/scenes at the same place.
  \item Reject if the query can be paraphrased as 'one photo of A, one of B, one of C at the same place'.
  \item Reject 'different aspects / different views / different angles / different corners' phrasing.
  \item A good query answers 'what changes or progresses within this place across the images?'.
  It must not merely answer 'what different things can be seen at this place?'.
\end{itemize}

CROSS-LOCATION FAMILY RULES:
\begin{itemize}[leftmargin=10pt]
    \item The defining logic must come from what STAYS STRUCTURALLY THE SAME across locations
  (repeated ritual / same semantic role / recurring activity / route-level progression).
  
    \item Reject checklists of destinations, attractions, landmarks, or 'different aspects of a trip'.

    \item Reject if the query can be paraphrased as 'one photo from place A, one from place B, one from place C'.

    \item cross.5 (generic shared pattern) is a last-resort template. Prefer cross.1/2/3/4 whenever
  a concrete ritual / role / activity / route applies. Do NOT use cross.5 for 'different nice streets/beaches/landmarks'.
  
    \item A good query answers 'what is being repeated or structurally preserved across places?'.
  It must not merely answer 'which places should appear?'.
\end{itemize}

GENERIC FAMILY RULES: Only accept when the bundle is held together by a true cross-image relation.
ALLOWED TEMPLATES for this candidate (choose exactly one template\_id):
\begin{itemize}[leftmargin=10pt]
  \item same.1: Same location + visible state or quantity change. The bundle must show a gradual, observable change within a fixed setting (street sparse->crowded, table empty->full->finished, campsite quiet->active).
  \item same.2: Same event in one place + stage progression. The bundle must move through distinct phases of one episode (setup -> peak -> aftermath; decorations -> celebration -> cleanup).
  \item same.3: Same place + before/after transition. Location identifiable, visible state change between earlier and later condition (before guests vs. during vs. after; before decorations lit vs. after).
  \item same.4: Same place + day-to-night / lighting progression. Location preserved, lighting/atmosphere visibly shifts over the day.
  \item same.5: Same place + complementary event summary. Tightly linked slices of ONE coherent episode in one place; each image plays a distinct but jointly-necessary role. Use this ONLY when progression is not monotonic but the slices are clearly one episode.
  
 \item cross.1: Repeated travel ritual across locations. Same personal ritual repeated in several different places (first meal after arrival, first hotel-window view, entrance-sign arrival photo).
\item cross.2: Same moment right after arrival across cities. A fixed temporal role in a repeated travel routine (first street scene after arrival, first check-in view, first stop after getting there).
  \item cross.3: Route-level progression across locations. Ordered movement through space during ONE continuous trip (hotel -> old town -> viewpoint; trailhead -> mid-route -> summit; city center -> waterfront -> harbor at sunset).
  \item cross.4: Repeated concrete activity across locations. Same specific visually-grounded act documented the same way in different places (breakfast-before-eating shots across cities, market-visit shots across trips, campsite-setup across camps).
  \item cross.5: Shared pattern stronger than location identity (generic fallback). Higher-level visual/semantic pattern stable across different places. Use ONLY when the shared pattern is specific, concrete, and strong enough to define a unique bundle. Do NOT use for 'different nice streets / beaches / landmarks'.

\end{itemize}

CALIBRATION EXAMPLES THAT MUST BE REJECTED:
\begin{itemize}[leftmargin=10pt]
    \item REJECT: "Find photos documenting a day trip in Mendoza, Argentina that includes relaxing at a hotel pool, viewing the hotel's glass architecture, walking to a stone-facade restaurant, dining with a group, and the hotel's Christmas-decorated entrance at night."  ->  decomposable checklist of 5 independent single-image targets; uses 'including ...'
  \item  REJECT: "Find images showing different aspects of a day trip to Santorini, Greece: the black sand beach, small boats on the Aegean Sea, traditional Cycladic architecture, and the cliffside town of Fira at dusk."  ->  tourism checklist; uses 'different aspects of'; each item is a separate single-image task
  \item  REJECT: "Find all photos from a single-day hiking expedition in the Andes mountains near Aconcagua, showing hikers on trail, mountain valley landscape, pack mules carrying supplies, and local wildlife."  ->  uses 'Find all photos'; the content is a 'showing A, B, C, and D' enumeration with no shared anchor
  \item  REJECT: "Find all photos from the same day showing this group's holiday gatherings at both the restaurant with ornate lantern lighting and the home with red walls and Christmas decorations in Mexico City."  ->  uses 'Find all photos'; blurs bundle boundary; two scene slices not tied by explicit progression\end{itemize}

Return STRICT JSON matching this schema (no markdown, no commentary):
\begin{Verbatim}[breaklines=true, breakanywhere=true]
    {
  "accept": "true or false",
  "query": "Natural English retrieval request, or null when rejected.",
  "image_ids": [photo_id_1, photo_id_2, ...],
  "rationale": {
    "template_id": "one of same.1..same.5 or cross.1..cross.5",
    "shared_anchor": "REQUIRED. A concrete, instance-level cross-image anchor that holds the bundle together. Must name a specific person / piece of equipment / landmark / ritual / repeated action / named route. MUST NOT be a generic phrase such as 'temporal progression', 'same day', 'single trip', 'the event', 'the route', 'cross-image reasoning'. Example: 'the same paddler wearing a blue hat, together with the same green canoe, appears in every image'.",
    "why_complete": "Why the set is only complete jointly. One sentence.",
    "why_unique": "Why this bundle is the most specific answer within the available pool.",
    "why_linked": "What cross-image relationship binds the images together. One sentence.",
    "why_not_decomposable": "Why this is not just multiple single-image tasks. One sentence.",
    "skeleton": "A/B/C"
  },
  "audit": {
    "single_image_insufficiency": "pass or fail with short justification",
    "non_substitutability": "pass or fail with short justification",
    "low_redundancy": "pass or fail with short justification",
    "natural_language_query": "pass or fail with short justification",
    "template_fit": "pass or fail, explain in <= 20 words why template_id applies",
    "review_risk": "low / medium / high"
  },
  "reject_reason": null
}
\end{Verbatim}

  \vspace{1em}

If rejecting, set accept=false, query=null, still fill rationale.template\_id with the closest template you considered, fill shared\_anchor with whatever concrete anchor (or 'none') you could identify, and put the specific failure in reject\_reason.
\end{tcolorbox}

\subsubsection{Details of Programmatic Gates}
\label{appendix: program gate}

The judging prompt in Appendix~\ref{appendix: verify prompt}  explicitly states the acceptance criteria
(forbidden query patterns, concrete anchor requirements, and
template-family constraints. Since the VLM does not always comply with
these instructions, we deterministically verify each accepted
output against the same rule set using regular expressions and
keyword matching. Specifically, we re-check: 
(1) the generated
query text for forbidden surface patterns, (2) the
\texttt{shared\_anchor} field for sufficient specificity (rejecting
anchors shorter than 4~tokens or dominated by generic phrases from a 28-term blocklist), and (3) the \texttt{template\_id} for membership in the candidate's allowed template family.
Any violation overrides the model's acceptance, converting it to a programmatic rejection.

\section{Implementation Details}
\label{appendix: implementation details}

All methods, including baselines and \model, are evaluated on the full \dataset dataset (\ie, 667 queries and bundles). For our proposed \model\ , we use RzenEmbed with a FAISS IndexFlatIP backend~\cite{douze2025faiss} for all base retrieval operations. We set the number of seed $N_s=5$, beam width $B_u=4$, and the candidates retrieved per step $C=5$. For candidate pruning,  we set the max bound to images captured within $\tau_t=24$ hours and $\tau_g=50$ km of the seed. The balancing coefficient $\lambda$ in Equation~\ref{eq:path scoring} is set to 0.3.  Unless otherwise specified, GPT-4o is used for both adaptive sub-query generation and whole-bundle pointwise reranking. All experiments are conducted on two A100 GPUs.

\subsection{Prompt Demonstration}
\subsubsection{Prompt for Adaptive Sub-Query Generation}

\begin{tcolorbox}[promptstyle, title=Pointwise Reranking]
\textbf{System:} You are assisting in constructing a photo bundle from a personal photo collection.

\textbf{User:} Original query: \{query\}

Previous search directions:

  1. \{sub\_query\_1\}
  
  2. \{sub\_query\_2\}
  
  ...

Images found so far:

  Image 1: \{caption\_1\}
  
  Image 2: \{caption\_2\}
  
  ...

Based on the original query and images already found, determine what is STILL MISSING from this photo bundle.
\begin{itemize}[leftmargin=10pt]
    \item If something is missing, output exactly one line: SEARCH: <concise description of the photo to find next>
    \item If the bundle appears complete (all aspects of the query are covered), output exactly: DONE
\end{itemize}

Output only SEARCH: ... or DONE, nothing else.
\end{tcolorbox}

\subsubsection{Prompts for Whole-Bundle VLM Reranking}

\begin{tcolorbox}[promptstyle, title=Pointwise Reranking]
\textbf{System:} You are evaluating whether a set of photos forms a coherent bundle matching a user query.

\textbf{User:} Query: \{query\}

This bundle has \{k\} photos:

\{Photo 1\}
  
\{Photo 2\}
  
...

Rate how well this photo bundle satisfies the query on a scale of 1-10:
\begin{itemize}[leftmargin=10pt]
    \item 10: Perfect match — every photo is relevant, together they fully satisfy the query.
    \item 7-9: Strong match — most photos are relevant, the bundle largely satisfies the query.
    \item 4-6: Partial match — some photos fit but the bundle is incomplete or has irrelevant photos.
    \item 1-3: Poor match — photos mostly don't relate to the query.
\end{itemize}

Output EXACTLY: SCORE: <number>

Then one sentence explaining why.

\end{tcolorbox}

\begin{tcolorbox}[promptstyle, title=Listwise Reranking]
\textbf{System}: You are a photo-album curator. Given a user query and \{n\} candidate photo bundles, pick the ONE bundle that best satisfies the query as a coherent set.

\textbf{User:} Query: \{query\}

\{candidates\}

For each bundle, ask:

\begin{itemize}[leftmargin=10pt]
    \item[1.] Do ALL photos belong to the SAME event/scene/trip described in the query?
    \item[2.] Does the set cover the key moments the query asks for?
    \item[3.] Are there irrelevant or duplicate photos?
    
\end{itemize}

Reply with EXACTLY one line:

BEST: <number>

Then one sentence explaining why.
\end{tcolorbox}

\subsection{Baseline Implementation}
\label{appendix:baseline_impl}

\paragraph{Multimodal Embedding.}
We encode the query text and all pool images independently using four vision-language embedding models: CLIP-ViT-B/32, SigLIP2-giant, Qwen3-VL-Embedding-8B, and RzenEmbed-7B.
All image embeddings are L2-normalized and indexed with FAISS \texttt{IndexFlatIP} for maximum inner product search, which is equivalent to cosine similarity for normalized vectors.
At query time, the query is encoded by the same multimodal embedding model.
This type of methods tests whether a single shared embedding space can capture the \emph{joint} semantics of a multi-image bundle from a text query alone.

\paragraph{Caption + Text Embedding.}
We first generate a detailed English caption for every image in the pool using GPT-4o, then perform text-to-text retrieval between the query and image captions. Three retrieval models are evaluated:
\begin{itemize}[leftmargin=10pt]
    \item \textbf{BM25}: a sparse lexical model using whitespace-tokenized captions.
    \item \textbf{BGE-M3}: using only the dense retrieval branch with FAISS indexing.
    \item \textbf{Qwen3-Embedding-8B}: a recent instruction-aware text embedding model where queries and passages are encoded with separate prompts.
\end{itemize}
Both dense models use L2-normalized embeddings with inner product search. This type of methods isolates whether richer textual descriptions of individual images can compensate for the lack of visual features in bridging the query--bundle gap.

\paragraph{Heuristic Metadata Augmentation.}
Building on the best multimodal model (RzenEmbed), we provide another line of baselines incorporating spatiotemporal metadata available in the photo manifest.
Three strategies are tested:

\begin{itemize}[leftmargin=10pt]

\item \textbf{Session Clustering:}
The retrieved images are sorted chronologically and partitioned into discrete sessions. A new cluster boundary is formed whenever consecutive images exhibit a time gap $> 6$ hours or a geographic jump $> 20$ km. The cluster with the highest aggregate embedding score is then selected, yielding its top-$|B^\ast|$ images. 

\item \textbf{User + Session Clusteing:}
The retrieved images are first grouped by photographer.
The dominant user is identified as the one whose top-$3|B^\ast|$ images have the highest aggregate embedding score, where $|B^\ast|$ denotes the size of the ground-truth bundle.
Session clustering is then applied within that user's images only. This leverages the observation that ground-truth bundles always belong to a single user.

\item \textbf{Time proximity boosting:}
For each retrieved image $i$, a bonus is added based on the embedding scores of temporally nearby images:
$\text{boost}_i = \sum_{j:\,|\Delta t_{ij}| \leq 1\text{h}} 0.1 \cdot s_j \cdot (1 - |\Delta t_{ij}|)$,
where $s_j$ is the original embedding score of image~$j$ and $|\Delta t_{ij}|$ is the time difference in hours. This linearly-decaying bonus encourages selecting images from the same short-duration event.

\item \textbf{User + Spatiotemporal:}
The retrieved candidates are first grouped by user. For each user, the image with the highest embedding score is designated as the anchor. The user's candidate pool is then explicitly filtered to retain only images falling within a $\pm 24$-hour window and a $50$\,km geographic radius relative to the anchor. The dominant user is identified by the highest aggregate score of their top-$3|B^\ast|$ filtered candidates, and their top-$|B^\ast|$ images are returned as the final bundle.

\end{itemize}
% \textit{Addtional Comment:} Session Clustering and User + Spatiotemporal baselines are specially introduced to investigate whether explicitly injecting the dataset construction priors or rigid metadata filtering could trivially solve IBC. Surprisingly, both heuristic augmentations degrade the set-level performance compared to the vanilla RzenEmbed model. This degradation highlights that naively enforcing spatiotemporal boundaries often truncates semantic continuity or forcibly groups physically close but relationally disjointed noise. It further suggests that IBC cannot be shortcut by simply recovering metadata priors. Instead, active inline reasoning  and whole-bundle coherence verification is indispensable.

\paragraph{VLM Decompose \& Rerank.}
This is the strongest baseline, employing a two-stage agentic pipeline.
In the \emph{decomposition} stage, a VLM is prompted to break the bundle query into $m$ sub-queries ($m \leq 5$), each describing one individual image expected in the bundle.
In the \emph{retrieval} stage, each sub-query is independently encoded using the RzenEmbed model and the top-5 most similar images are retrieved from the FAISS index, yielding up to $5m$ candidates.
In the \emph{reranking} stage, the VLM receives all candidate images alongside the original query and is prompted to select exactly one image per sub-query group.
If the VLM fails to parse or produces invalid selections, a fallback selects the top-1 retrieval result from each sub-query group. We evaluate five VLMs: GPT-4o, Claude-Sonnet-4.5, Gemini-3-Flash-Preview, Qwen2.5-VL-72B, and Qwen3-VL-235B.
Note that the number of selected images~$m$ is determined by the VLM's decomposition and may differ from the ground-truth bundle size~$|B^\ast|$.

\section{Additional Experiments}
\label{appendix: more experiments}

\subsection{Efficiency Analysis}
\label{appendix: efficiency}

We provide a detailed efficiency analysis of BundleWeaver and compare it with representative retrieval baselines. All experiments are conducted on a 109K-image pool with precomputed image captions and embeddings. For standard embedding-based retrieval, rzenembed requires only 0.3s per query, while metadata-augmented retrieval with Session Clustering takes 0.8s. The static decompose-and-rerank agentic baseline based on GPT-4o takes 12s per query. 

In contrast to these atomic retrieval approaches, BundleWeaver adopts an agentic retrieval paradigm that performs iterative reasoning, targeted search, and verification to solve the more challenging IBC task. The latency breakdown of BundleWeaver is shown in Table~\ref{tab:efficiency}. The additional latency of BundleWeaver mainly comes from its multi-stage agentic reasoning and verification process. This trade-off is expected: agentic retrieval replaces one-shot embedding matching with iterative exploration and bundle-level reasoning. Similar latency challenges are common in agentic retrieval systems. Nevertheless, our objective is not to provide a low-latency alternative to conventional embedding search, but rather to provide the first effective solution for the significantly harder IBC setting, where atomic retrieval methods are fundamentally insufficient.

\begin{table}[t]
\centering
\small
\begin{tabular}{lc}
\toprule
\textbf{Component} & \textbf{Latency} \\
\midrule
LLM anchor extraction + seed initialization & 3s \\
Global ANN retrieval + diverse seed selection & 1s \\
Iterative missing-role subquery generation & 10s \\
Local retrieval + beam expansion & 3s \\
Whole-bundle VLM reranking & 13s \\
System overhead & 2s \\
\midrule
Total & 32s \\
\bottomrule
\end{tabular}
\caption{Latency breakdown of BundleWeaver on a 109K-image pool.}
\label{tab:efficiency}
\end{table}

\begin{table}[t]

\centering

% \vspace{-10pt}
% \footnotesize
\resizebox{0.47\textwidth}{!}{
% \renewcommand\arraystretch{1.1}

% \usepackage[normalem]{ulem}
\begin{tabular}{c|cccc}
\toprule
\hline
Strategy  & Precision      & Recall         & F1             & EM            \\ \hline
No-rerank & 26.33          & 26.70          & 26.13          & 6.15          \\
Listwise  & \underline{ 29.41}    & \textbf{31.28} & \underline{ 29.94}    & \underline{ 6.90}    \\
Pointwise & \textbf{30.95} & \underline{ 30.46}    & \textbf{30.28} & \textbf{7.20} \\ \hline

\bottomrule

\end{tabular}
}
\caption{Performance comparison of different reranking strategies.
}
\label{tab:rerank}

    \vspace{-3pt}

\end{table}

\subsection{Reranking Strategy Comparison}

For reranking strategy in Section~\ref{sec: vlm rerank}, we compare \textit{Pointwise} scoring against \textit{Listwise} selection (\eg, asking the VLM to pick the best from all constructed bundles). From Table~\ref{tab:rerank}, we can see that rerank generally helps, and pointwise evaluation is more rigorous and scalable. While Listwise evaluation achieves competitive Recall, Pointwise scoring yields the best F1 and EM. We attribute this to attention dilution in Listwise prompts: when faced with dozens of images simultaneously, VLMs struggle to verify intricate constraints, which is consistent with previous works~\cite{meng2025mmiu, lyu2025vischainbench}. Pointwise scoring forces the VLM to independently and rigorously audit each bundle, also offering better scalability than Listwise reranking.

\subsection{Expanded Baseline Evaluation}
For non-agentic methods, oracle-size truncation may assume prior knowledge of the ground-truth bundle cardinality. However, this evaluation protocol provides retrieval baselines with a strong oracle prior: for each query, the ranked retrieval list is truncated to exactly $|B^*|$, removing the need for baselines to determine how many images should be returned. This setting is favorable to atomic retrieval methods, yet they still substantially underperform BundleWeaver.

To further examine this issue, we introduce a dynamic-cardinality evaluation setting. Instead of using the ground-truth bundle size, retrieval methods return all images whose similarity scores exceed a predefined threshold. Results with RzenEmbed are shown in Table~\ref{tab:dynamic_cardinality}. Without oracle cardinality information, atomic retrieval becomes less effective: high thresholds improve precision but miss many required bundle members, while low thresholds increase recall at the cost of introducing visually similar distractors.

\begin{table}[t]
\centering

\resizebox{0.47\textwidth}{!}{
\begin{tabular}{c|c|ccc}
\toprule
\hline
Method & Truncation Rule & Precision & Recall & F1 \\
\hline

\multirow{5}{*}{RzenEmbed}
& Oracle size & 14.88 & 14.88 & 14.88 \\
& Score $>$ 0.95 & 18.2 & 3.7 & 6.1 \\
& Score $>$ 0.90 & 11.8 & 8.2 & 9.7 \\
& Score $>$ 0.85 & 7.4 & 13.6 & 9.6 \\
& Score $>$ 0.80 & 4.1 & 18.9 & 6.7 \\
\hline

BundleWeaver (Ours)
& N/A & \textbf{30.95} & \textbf{30.46} & \textbf{30.28} \\

\hline
\bottomrule

\end{tabular}
}

\caption{Dynamic-cardinality evaluation compared with oracle-size truncation.}
\label{tab:dynamic_cardinality}

\vspace{-3pt}

\end{table}

Removing the oracle cardinality prior makes atomic retrieval even less suitable for IBC: strict thresholds fail to recover complete bundles, while loose thresholds introduce many irrelevant candidates. Therefore, our original oracle-size evaluation is conservative and favors retrieval baselines. The performance gap between atomic retrieval and BundleWeaver is expected to be even larger under realistic dynamic-cardinality settings.

\subsection{Hyperparameter Study}

\begin{figure}[t]
    \centering
    \includegraphics[width=0.99\linewidth]{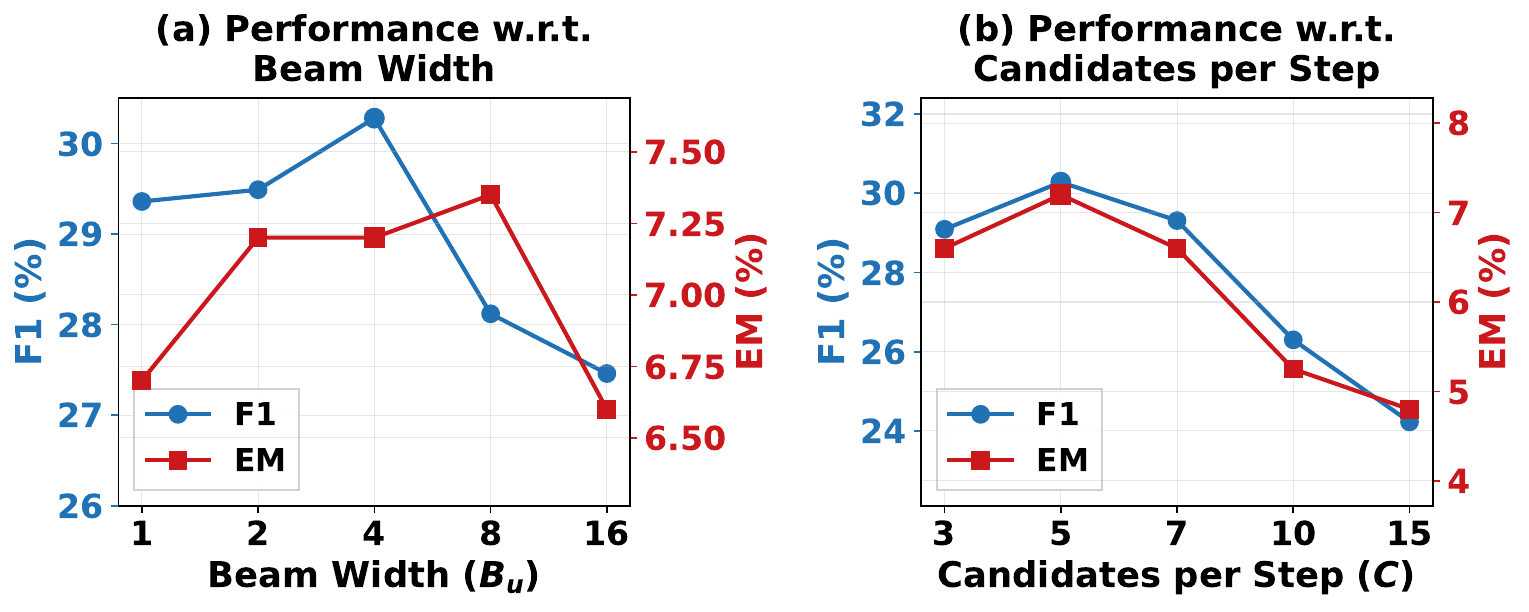}
    \caption{Hyperparameter study of the beam width and candidates per step.}
    \label{fig:hyper study}
\end{figure}

\begin{figure}[t]
    \centering
    \includegraphics[width=0.99\linewidth]{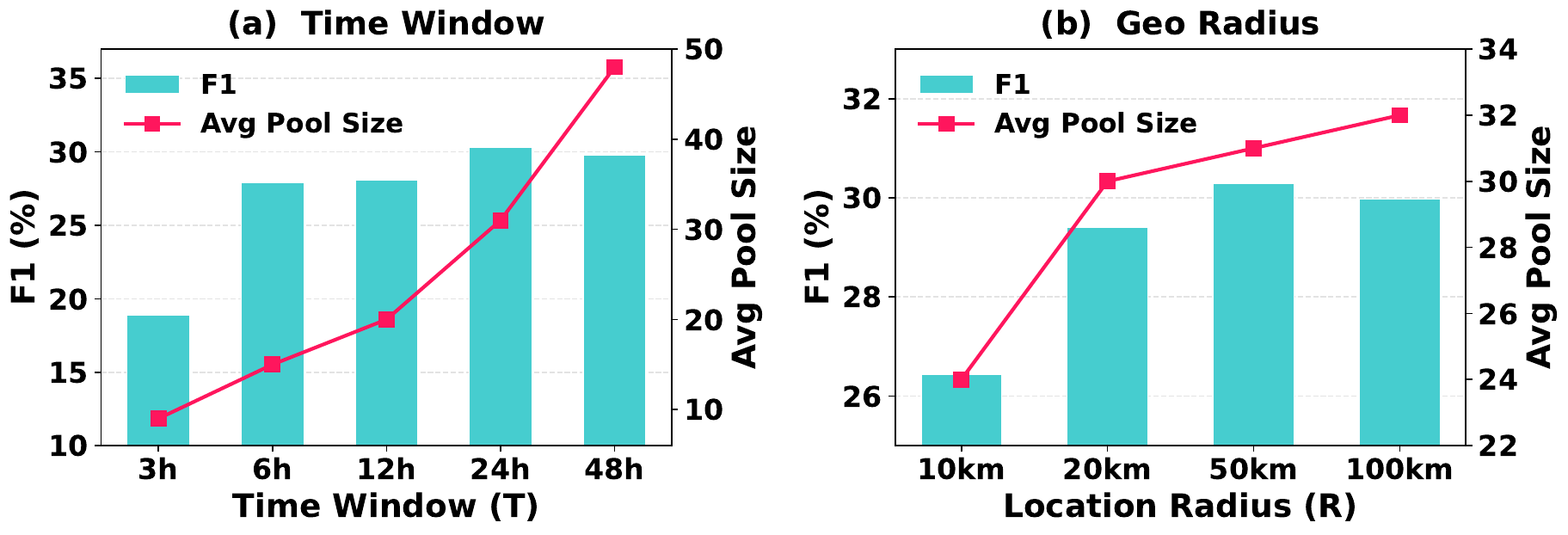}
    \caption{Hyperparameter study of the candidate pruning range.}
    \label{fig:metadata}
\end{figure}

\paragraph{Parameters of Beam Search.} In Figure~\ref{fig:hyper study}, we investigate the sensitivity of \model\ to its core exploration hyperparameters: the parallel beam width ($B_u$) and the number of local candidates retrieved per step ($C$).  Initially scaling up these parameters improves performance by providing the reasoning agent with a richer, more diverse set of visual building blocks. However, over-expanding the beam width or retrieving an excessive number of candidates per step actively degrades the final retrieval quality. This occurs because unbounded exploration floods the candidate pool with noisy, locally similar but globally disjointed images, which ultimately derails the LLM's adaptive reasoning trajectory. The existence of a clear optimal (\ie, $B_u=4, C=5$) empirically indicates that solving IBC requires a delicate balance: while diverse exploration is necessary to avoid local optima, aggressive and strict pruning is equally vital to maintain a coherent narrative direction within a massive combinatorial space.

\paragraph{Parameters of Candidate Pruning.} 
We study the effect of the temporal window $T$ and geographic radius $R$ used by the agent in contextual candidate pruning. Figure~\ref{fig:metadata} reports F1 together with the average local pool size induced by each constraint. The pool size is computed with the first ground-truth image as the anchor only for diagnostic analysis, and this oracle anchor is not used by the retrieval method.

As shown in the figure, an overly tight temporal window ($T=3$h) substantially hurts performance, since it excludes valid bundles spanning longer events or day-long trips. Increasing $T$ to $12$--$24$h yields the best performance, while further enlarging it to $48$h increases the candidate pool without improving F1.  A similar trend can be observed for location: $R=10$km is too restrictive for cross-location bundles, whereas performance remains stable from $20$km to $100$km. These results show that our default setting ($T=24$h, $R=50$km)  maintains performance while keeping the candidate pool compact.

\subsection{Human Validation of Model Results}
\label{appendix: human validation}

\begin{table}[t]

\centering

% \footnotesize
\resizebox{0.48\textwidth}{!}{
\renewcommand\arraystretch{1.1}

\begin{tabular}{c|cccc}
\toprule
\hline
Method & Precision & Recall & F1 & \makecell{False  Negative \\ Rate} \\ \hline
Claude-Sonnet-4.5 & 22.20 & 21.48 & 21.66 & 0\% \\
\model\ & 26.27 & 24.47 & 25.08 & 2\% \\ \hline
\bottomrule
\end{tabular}
}

\caption{Manual audit of strict false negatives on 100 randomly sampled queries, where both \model\ and Claude-Sonnet-4.5 fail under automatic set-level evaluation.
}
\label{tab:human validation}
\end{table}

To assess whether strict set matching unfairly penalizes methods due to missing alternative valid bundles, we manually audit 100 randomly sampled queries where both \model\ and Claude-Sonnet-4.5 are marked incorrect by automatic exact set-level evaluation. As shown in Table~\ref{tab:human validation}, unannotated valid alternatives are rare (required by C3 in Section~\ref{sec: bundle requirements}): only 2\% of \model\ predictions are judged as reasonable alternative bundles, while none of the Claude-Sonnet-4.5 predictions are human-valid. This suggests that incomplete ground-truth annotation does not substantially distort our strict evaluation protocol. Meanwhile, even when both methods fail to recover the complete target bundle, \model\ achieves higher Precision, Recall, and F1, indicating better partial recovery of the ground-truth bundle images in difficult failure cases.

\subsection{Case Study}

\begin{figure*}[t]
    \centering
    \includegraphics[width=0.97\textwidth]{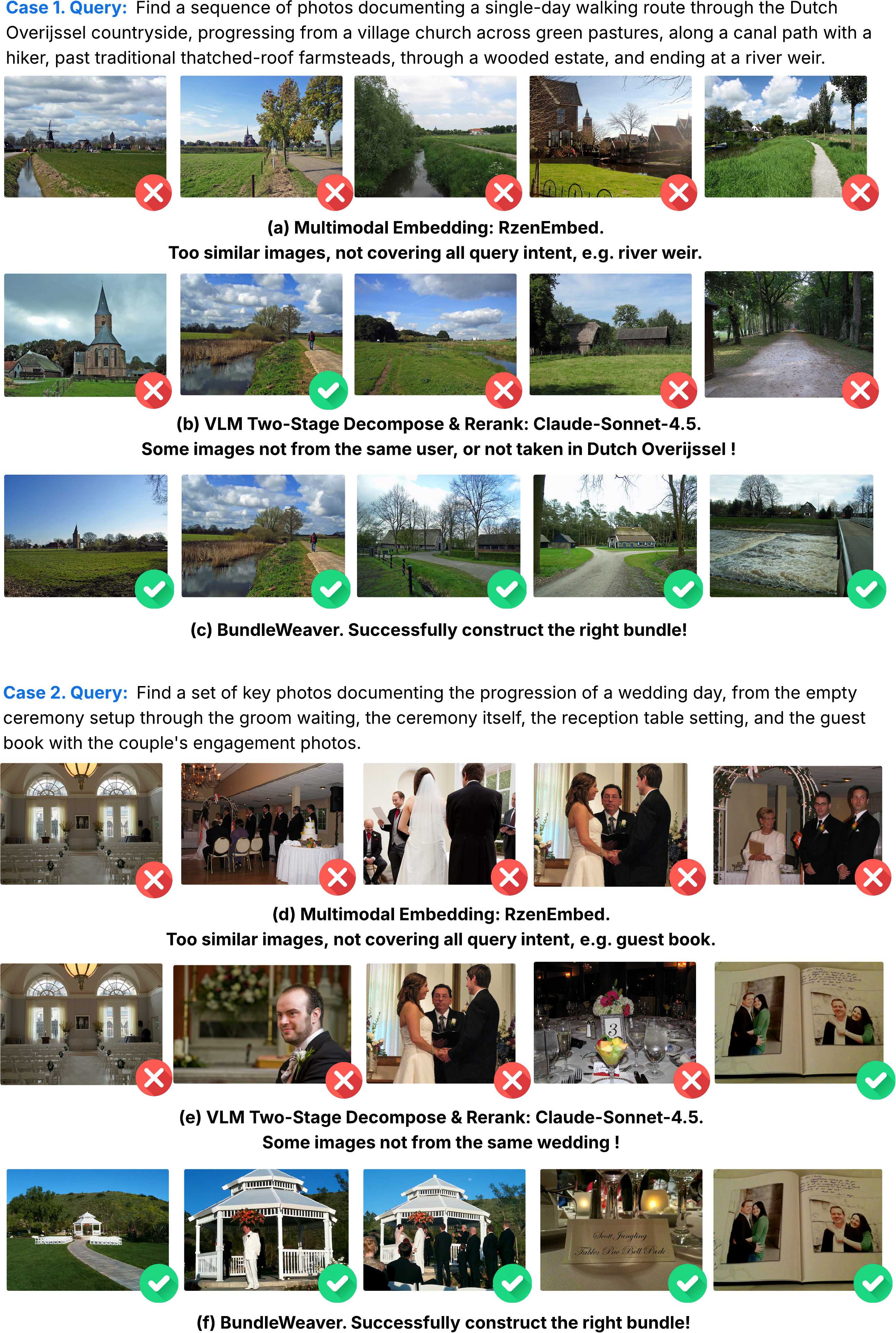}
    \caption{Examples showing the results of different methods on IBC.}
    \label{fig:cases}
\end{figure*}

To intuitively demonstrate the limitations of existing retrieval paradigms and the superiority of \model, we provide case studies on two representative queries from \dataset, as shown in Figure~\ref{fig:cases}.

\noindent\textbf{Case 1: Spatial Route Progression.} The query requests a continuous spatial narrative: a single-day walking route through the Dutch countryside, explicitly requiring a progression through distinct visual landmarks (a church, green pastures, a canal path, farmsteads, a wooded estate, and a river weir). 
\begin{itemize}[leftmargin=10pt]
    \item \textbf{Multimodal Embedding (RzenEmbed):} The traditional embedding model completely fails to capture the narrative progression. Instead, it suffers from \textit{visual collapse}, aggressively retrieving images that are visually homogeneous (\eg, predominantly green canal paths). This perfectly illustrates the flaw of atomic point-wise matching: it optimizes for isolated semantic similarity but ignores the sequential completeness of the query.
    \item \textbf{VLM Decompose \& Rerank (Claude-Sonnet-4.5):} The agentic decomposition baseline successfully identifies the distinct landmarks by breaking the query into independent sub-queries. However, because these sub-queries are retrieved in isolation, the resulting bundle is a \textit{Frankenstein} composition. The images belong to different trips, different users, or entirely different geographical regions, severely violating the real-world route continuity constraint required by IBC.
    \item \textbf{\model\ (Ours):} By anchoring the search within valid spatiotemporal metadata windows and using an LLM to adaptively search for the \textit{next missing landmark} conditioned on the current path, our method successfully constructs a coherent, authentic walking route that perfectly matches the user's compositional intent.
\end{itemize}

\noindent\textbf{Case 2: Temporal Event Progression.}
The query demands a temporal progression of a specific event: a wedding day, spanning from the empty setup and the waiting groom to the ceremony, reception table, and guest book. This query  requires \textit{instance-level identity consistency} (\ie, all photos must belong to the exact same wedding).
\begin{itemize}[leftmargin=10pt]
    \item \textbf{Multimodal Embedding (RzenEmbed):} The embedding model retrieves a redundant set of highly similar ceremony shots. The strong overarching semantic signal of \textit{wedding} overshadows the fine-grained requirements (like the empty setup or guest book), showing that dense embeddings struggle to compose diverse narrative slices.
    \item \textbf{VLM Decompose \& Rerank (Claude-Sonnet-4.5):} This baseline successfully retrieves the diverse semantic components (the groom, the ceremony, the table setting, the guest book). However, a closer inspection reveals a fatal flaw: \textbf{the images depict completely different couples and different weddings}. This is the ultimate proof of the "Relational Blindness" discussed in Section 5. The static divide-and-conquer strategy cannot enforce the implicit constraint that the core subject must remain identical across the independently retrieved subsets.
    \item \textbf{\model\ (Ours):} Our framework excels here. By incrementally expanding the bundle and utilizing pointwise VLM reranking to verify the whole-bundle logic, \model\ seamlessly tracks the temporal progression while maintaining strict identity consistency. It retrieves the exact timeline of a single, unique wedding day, showcasing its robust capability in modeling complex, non-decomposable joint relevance.
\end{itemize}

% \begin{figure*}[h]
%     \centering
%     \includegraphics[width=0.97\textwidth]{figures/case_02_70408381@N00_s8420_w0_n5/case2.png}
%     \caption{The second example showing the results of different methods on IBC.}
%     \label{fig:case2}
% \end{figure*}

\subsection{Failure Case Analysis}

While BundleWeaver substantially improves IBC retrieval performance by explicitly modeling relational completeness and bundle-level consistency, it can still fail in several challenging scenarios. We conduct a qualitative failure analysis and summarize four representative failure modes.
\begin{itemize}[leftmargin=10pt]
    \item \textbf{Incomplete relational coverage.}
A common failure occurs when the retrieved bundle contains visually relevant images but fails to cover all required roles in the underlying event. This typically happens when certain roles correspond to rare visual patterns, weakly observable objects, or images with limited semantic cues. Although BundleWeaver can generate missing-role subqueries, the generated queries may still be insufficient when the required role is absent from the candidate pool or difficult to distinguish from visually similar distractors.

\item \textbf{Identity and event inconsistency.}
BundleWeaver may retrieve individually relevant images that satisfy local subqueries but violate the global event-level constraint. For example, images captured by different users or from different occasions may share similar visual contexts, objects, or scenes. Without sufficient identity cues, the model may incorrectly combine these images into a single bundle, resulting in an inconsistent narrative.

\item \textbf{Broken temporal or narrative order.}
Another failure mode involves incorrect ordering of retrieved images. The model may identify semantically related images but fail to recover the intended temporal progression. This can lead to skipped intermediate stages, insertion of irrelevant distractors, or reversed event transitions. Such failures are particularly challenging because temporal relationships are often weakly represented in individual image embeddings.

\item \textbf{Metadata-neighborhood distractors.}
Although metadata information can effectively reduce the search space, metadata proximity does not necessarily imply relational relevance. Images that are physically close in time or location may belong to unrelated activities or different events. As a result, metadata-based retrieval can introduce misleading candidates, demonstrating that metadata serves as a useful auxiliary signal but cannot independently resolve the IBC problem.
\end{itemize}
Overall, these failure cases highlight the intrinsic difficulty of IBC retrieval: successful retrieval requires not only visual similarity, but also comprehensive role coverage, global consistency, and coherent event-level reasoning. Future improvements may focus on stronger identity-aware representations, more reliable temporal modeling, and adaptive verification strategies for ambiguous bundles.

\section{Discussions on Potential Concerns}
\label{appendix: response}

In this section, we address potential questions regarding our paper.

\subsection{Is contextual candidate pruning a benchmark bias?}

\begin{table}[t]

\centering

% \footnotesize
\resizebox{0.47\textwidth}{!}{
% \renewcommand\arraystretch{1.1}

% \usepackage[normalem]{ulem}
\begin{tabular}{c|cccc}
\toprule
\hline
Method  & Precision      & Recall         & F1             & EM            \\ \hline
RzenEmbed  & 14.88    & 14.88 & 14.88    & 0.30    \\
w/ CCP & 8.80 & 8.80    & 8.80 & 0.75 \\ \hline

GPT-4o Dec. \& Re. & 18.50 & 17.34 & 17.84 & 0.60 \\
w/ CCP & 16.75 & 16.74 & 16.74 & 3.15 \\ \hline

\model\ w/o CCP & \underline{24.57} & \underline{24.74} & \underline{24.69} & \underline{5.85} \\
\model & \textbf{30.95} & \textbf{30.46} & \textbf{30.28} & \textbf{7.20} \\ \hline

\bottomrule

\end{tabular}
}

\caption{Controlled experiment on contextual candidate pruning (CCP).
}
\label{tab: prune}

    \vspace{-3pt}

\end{table}

A potential concern is that \dataset{} is mined from spatiotemporal sessions, while \model{} also uses contextual candidate pruning (CCP). This raises the question of whether \model{} benefits from a dataset-specific shortcut rather than solving the intended relational composition problem.

We argue that CCP should be viewed as a \textit{query-faithful physical plausibility constraint} to ensure bundle uniqueness, rather than an artificial benchmark bias. Many IBC queries explicitly or implicitly contain temporal and geographic constraints, such as a single trip, a same-day event progression, a route across landmarks, or a repeated activity at nearby venues. In such cases, using timestamps and GPS metadata is part of faithfully executing the user intent, analogous to filtering by date or location in real-world personal photo search. Ignoring such metadata would create an unrealistic retrieval setting in which the model is forced to search globally even when the query itself specifies a physical context.

Importantly, however, CCP only defines a broad feasible candidate region; it does not determine the answer. Physical proximity is neither sufficient nor monotonically beneficial for IBC. As shown in Table~\ref{tab: prune}, adding the same CCP mechanism to RzenEmbed decreases F1 from 14.88 to 8.80, and adding it to GPT-4o Decompose-and-Rerank decreases F1 from 17.84 to 16.74, although EM improves slightly. This indicates that spatiotemporal locality may help recover some complete local bundles, but it also introduces many physically close yet relationally irrelevant distractors and can hurt partial set recovery.

This observation is also \textit{consistent} with our dataset construction process. Although candidate windows are first mined under spatiotemporal constraints, fewer than 9\% survive the subsequent semantic verification and human review. Thus, most spatiotemporally plausible windows are not valid IBC answers. The benchmark therefore does not reward recovering metadata priors alone; it requires identifying which images jointly instantiate the non-decomposable relation specified by the query.

Finally, \model{} remains strong even when CCP is removed. The w/o-CCP variant achieves 24.69 F1, substantially outperforming the strongest GPT-4o static baseline at 17.84 F1. This shows that the core gain of \model{} comes from adaptive missing-role reasoning, beam-based bundle construction, and whole-bundle verification. CCP mainly serves as a practical physical boundary that reduces global combinatorial noise, rather than as a shortcut to the ground-truth bundle.

\subsection{Is the "uniqueness" assumption fully validated?}

Ground-truth uniqueness leads to robust evaluation. IBC evaluation relies on exact set-matching, which assumes a single ground-truth bundle per query. A critical question is whether bundles annotated in \dataset\ have multiple \textit{reasonable alternative} answers, which would cause our evaluation to unfairly penalize valid predictions. 

To address this, we highlight the criteria for uniqueness and provide quantified evidence that false negatives (penalizing a valid alternative) are minimized.

\paragraph{Quantified Error Audit.} As reported in Table~\ref{tab:human validation}, we manually inspected 100 randomly sampled errors where both \model\ and Claude-Sonnet-4.5 failed the automatic evaluation. A predicted bundle is judged as a \textit{reasonable alternative} (\ie, a genuine false negative) \textbf{ONLY IF} it fully satisfies constraints C1-C4 described in Section~\ref{sec:dataset_construction} and perfectly aligns with the text query, but consists of different images than the ground truth. Upon rigorous inspection, the false negative rate was merely 2\% for \model\ and 0\% for Claude. In $>98\%$ of the failure cases, the predicted bundles were genuinely flawed (\eg, containing irrelevant distractors, breaking the narrative timeline, or violating identity consistency).

\paragraph{How Uniqueness is Enforced During Construction.} The low false negative rate is a direct result of our strict human verification design (C3). During annotation, reviewers were equipped with a Nearest-Neighbor visual panel (Figure~\ref{fig: review web}) to explicitly hunt for alternative bundles.
\begin{itemize}[leftmargin=10pt]
    \item \textbf{Example of a rejected query:} If a VLM generates \textit{"Find 3 photos of the couple kissing at the wedding"}, and the pool contains a burst of 10 kissing photos, any subset of 3 would be a valid alternative. Because it violates uniqueness, \textbf{the annotator rejects this query entirely}. 
    \item \textbf{Example of an accepted query:} The annotator refines the query to "Find a progression of the ceremony: from the empty setup, to the groom waiting, and finally the kiss." This structural constraint forces a unique, unambiguous trajectory, eliminating reasonable alternatives.
\end{itemize}
By explicitly rejecting any queries with interchangeable subset combinations during construction, we try our best to ensure that the exact set-matching evaluation metric remains highly reliable and fair.

\section{Theoretical Limitations of Atomic Retrieval in IBC}
\label{appendix:appendix_theoretical_limitations}

In this section, we mathematically demonstrate why the static Decompose-and-Rerank paradigm inherently suffers from \textit{Relational Blindness} when applied to Image Bundle Composition (IBC). We first prove the non-submodularity of the IBC objective, which subsequently establishes the theoretical bound where local top-$k$ retrieval fails to converge to the global optimum.

\subsection{Preliminaries and Problem Definition}

As defined in Equation~\ref{eq: BIR objective}, the objective of IBC is to find the optimal image subset $B^*$ that maximizes the joint relevance score:
\begin{equation*}
\label{eq:bi_objective}
B^* = \arg\max_{\substack{B \subset \mathcal{I} ,  2 \le |B| \le K_{max}}} \Phi(B, q)
\end{equation*}

In the Decompose-and-Rerank paradigm, a complex query $q$ is factorized into $K$ independent sub-queries $\mathcal{Q} = \{q_1, q_2, \dots, q_K\}$. The joint relevance scoring function $\Phi(B, q)$ can be conceptually formulated as the sum of local semantic matching scores and a global cross-image relational constraint:
\begin{equation}
\label{eq:joint_relevance}
\Phi(B, q) = \sum_{i=1}^K f(x_i, q_i) + \gamma \cdot \mathbb{I}_{rel}(B),
\end{equation}
where:
\begin{itemize}[leftmargin=10pt]
    \item $f(x_i, q_i)$ denotes the independent visual-textual alignment score between a single candidate image $x_i$ and the sub-query $q_i$.
    \item $\mathbb{I}_{rel}(B) \in \{0, 1\}$ is a non-decomposable Boolean indicator function representing whether the subset $B$ satisfies the global relational constraint dictated by $q$ (\eg, spatiotemporal continuity, identity consistency).
    \item $\gamma \to \infty$ represents a strict hard constraint penalty (\ie, a bundle that breaks the relational logic is fundamentally invalid).
\end{itemize}

\subsection{Lemma 1: Strict Non-Submodularity of Relational Bundles}

In traditional retrieval and subset selection problems, the scoring function is often assumed to be submodular, allowing greedy algorithms to yield bounded approximation guarantees. We demonstrate that IBC violates this foundational assumption.

\vspace{0.5em}
\noindent \textbf{Definition (Submodularity).} Let $V$ be a finite set. A set function $\Phi: 2^V \to \mathbb{R}$ is submodular if and only if for every $A \subseteq B \subset V$ and $x \notin B$, it satisfies the diminishing returns property:
\begin{equation}
\label{eq:submodularity}
\begin{split}
&\Phi(A \cup \{x\}) - \Phi(A) \\
&\ge \Phi(B \cup \{x\}) - \Phi(B)
\end{split}
\end{equation}

\vspace{0.5em}
\noindent \textbf{Lemma 1.} \textit{The joint relevance scoring function $\Phi(B, q)$ for IBC is strictly non-submodular when subject to relational structural constraints.}

\vspace{0.5em}
\noindent \textit{Proof.} We prove this by construction. Consider a temporal event progression query $q$ (\eg, the wedding progression in Figure~\ref{fig:cases}). Let the target optimal bundle consist of three chronological stages: $B^* = \{x_{setup}^*, x_{wait}^*, x_{ceremony}^*\}$. 

Let subset $A = \{x_{setup}^*\}$ and subset $B = \{x_{setup}^*, x_{wait}^*\}$, where $A \subset B$. We now introduce a new candidate image $x_{ceremony}^*$ (a ceremony shot from the exact same wedding).

\begin{enumerate}
    \item When $x_{ceremony}^*$ is added to $A$, the subset becomes $\{x_{setup}^*, x_{ceremony}^*\}$. Because the intermediate temporal bridge ($x_{wait}^*$) is missing, the narrative progression is broken, yielding $\mathbb{I}_{rel}(A \cup \{x_{ceremony}^*\}) = 0$. The marginal gain is strictly the local semantic score:
\begin{equation}
\label{eq:marginal_gain_A}
\begin{split}
&\Phi(A \cup \{x_{ceremony}^*\}) - \Phi(A) \\
&= f(x_{ceremony}^*, q_{ceremony})
\end{split}
\end{equation}

    \item When $x_{ceremony}^*$ is added to $B$, the subset becomes the complete chronological story $\{x_{setup}^*, x_{wait}^*, x_{ceremony}^*\}$. The narrative is successfully closed, satisfying the joint constraint $\mathbb{I}_{rel}(B \cup \{x_{ceremony}^*\}) = 1$. The marginal gain is:
\begin{equation}
\label{eq:marginal_gain_B}
\begin{split}
&\Phi(B \cup \{x_{ceremony}^*\}) - \Phi(B) \\
&= f(x_{ceremony}^*, q_{ceremony}) + \gamma
\end{split}
\end{equation}

Since $\gamma > 0$ denotes the massive reward for satisfying the hard relational constraint, we obtain:
\begin{equation}
\label{eq:non_submodular_result}
\begin{split}
&\Phi(A \cup \{x_{ceremony}^*\}) - \Phi(A) \\
&< \Phi(B \cup \{x_{ceremony}^*\}) - \Phi(B)
\end{split}
\end{equation}

This strict inequality violates the diminishing returns property of submodular functions. $\hfill \blacksquare$
\end{enumerate}

\vspace{0.5em}
\noindent \textbf{Corollary.} Since $\Phi$ is strictly non-submodular, any greedy selection algorithm relying on independent marginal gains cannot guarantee a constant-factor approximation bound. This necessitates an adaptive, context-aware expansion strategy like BundleSearch.

\subsection{Theorem 1: The Relational Blindness Bound of Independent Top-$k$}

Building upon Lemma 1, we formally establish the failure mechanism of the Decompose-and-Rerank paradigm. In this paradigm, the system independently retrieves the top-$k$ candidates for each sub-query $q_i$, forming a local candidate space $\mathcal{S}_i = \arg\max_{S \subset \mathcal{I}, |S|=k} \sum_{x \in S} f(x, q_i)$.

\vspace{0.5em}
\noindent \textbf{Theorem 1.} \textit{Assume that for any target image $x_i^* \in B^*$, there exist $M$ ``Dominant Distractors'' $x_{i, d}$ in the massive unindexed image pool $\mathcal{I}$. These distractors satisfy:}
\begin{enumerate}
    \item \textit{Higher local semantic alignment: $f(x_{i, d}, q_i) > f(x_i^*, q_i)$.}
    \item \textit{Violation of global relational constraints: For any bundle $B'$ containing $x_{i, d}$, $\mathbb{I}_{rel}(B') = 0$.}
\end{enumerate}
\textit{If the retrieval system's cutoff parameter satisfies $k \le M$, the probability $P(B^*)$ of successfully discovering the optimal bundle $B^*$ via the Decompose-and-Rerank paradigm is strictly 0.}

\vspace{0.5em}
\noindent \textit{Proof.} During the independent retrieval stage, the global constraint $\mathbb{I}_{rel}$ is invisible to the atomic scorer. Because there are $M$ dominant distractors satisfying $f(x_{i, d}, q_i) > f(x_i^*, q_i)$, the greedy top-$k$ selection will exclusively populate $\mathcal{S}_i$ with these distractors (since $k \le M$). Consequently, the true target image is deterministically excluded from the candidate pool: $x_i^* \notin \mathcal{S}_i, \forall i \in \{1, \dots, K\}$.

During the selection stage, the VLM is restricted to search within the Cartesian product $\mathcal{S}_{global} = \mathcal{S}_1 \times \mathcal{S}_2 \times \dots \times \mathcal{S}_K$. Since every candidate bundle $B_{cand} \in \mathcal{S}_{global}$ contains at least one dominant distractor $x_{i,d}$, it follows from our assumption that:
\begin{equation}
\label{eq:zero_relational_constraint}
\forall B_{cand} \in \mathcal{S}_{global}, \quad \mathbb{I}_{rel}(B_{cand}) = 0
\end{equation}

Therefore, the system is mathematically forced to output a fragmented composite with a joint relevance score significantly lower than $\Phi(B^*, q)$. Unless the cutoff $k$ is expanded such that $k > M$ (which causes the VLM's combinatorial search space $\mathcal{O}(k^K)$ to explode exponentially, rendering it computationally intractable), the independent splitting strategy will inevitably succumb to Relational Blindness. $\hfill \blacksquare$

\vspace{0.5em}
\noindent \textbf{Remark on Empirical Observations.} Theorem 1 directly explains the catastrophic drop in Exact Match (EM) rates for advanced VLMs reported in Table~\ref{tab:main_table}. As vividly illustrated in Case Study 2 (Figure~\ref{fig:cases}), visually stunning wedding photos from \textit{different} events act precisely as \textit{Dominant Distractors} ($x_{i, d}$). They easily overwhelm the local top-$k$ ranking due to high semantic similarity to the sub-query, entirely displacing the authentic, identity-consistent images ($x_i^*$) required to fulfill the global relational constraint.

\end{document}